\pdfoutput=1

\documentclass[11pt]{article}

\usepackage[final]{acl}

\usepackage{times}
\usepackage{latexsym}
\usepackage{amsmath}
\usepackage{algorithm}
\usepackage{algpseudocode}

\usepackage{promptbox}

\usepackage[normalem]{ulem}

\usepackage{multirow}
\usepackage{arydshln}
\usepackage{subcaption}
\usepackage{siunitx}
\usepackage{tcolorbox}

\usepackage{hyperref}

\usepackage[T1]{fontenc}

\usepackage[utf8]{inputenc}

\usepackage{microtype}

\usepackage{inconsolata}

\usepackage{graphicx}

\usepackage{xcolor}
\usepackage{colortbl}
\usepackage{booktabs} 
\usepackage{array}    

\definecolor{headergray}{gray}{0.9}
\definecolor{red2}{RGB}{210, 51, 31}

\newcommand\bluesout{\bgroup\markoverwith{\textcolor{blue}{\rule[0.5ex]{2pt}{0.8pt}}}\ULon}

\title{Query-driven Document-level Scientific Evidence Extraction from Biomedical Studies}

\usepackage{hyperref} 

\author{
 \textbf{Massimiliano Pronesti\textsuperscript{1,2}},
 \textbf{Joao Bettencourt-Silva\textsuperscript{1}},
 \textbf{Paul Flanagan\textsuperscript{2}},
 \textbf{Alessandra Pascale\textsuperscript{1}},
\\
 \textbf{Oisín Redmond\textsuperscript{2}},
 \textbf{Anya Belz\textsuperscript{2}\textsuperscript{$\dagger$}},
 \textbf{Yufang Hou\textsuperscript{1,3}\textsuperscript{$\dagger$}}
\\
 \textsuperscript{1}IBM Research Europe - Ireland,
 \textsuperscript{2}Dublin City University,\\
 \textsuperscript{3}IT:U Interdisciplinary Transformation University Austria
\\
 \small{
   \textbf{Correspondence:} \href{mailto:massimiliano.pronesti@ibm.com}{massimiliano.pronesti@ibm.com}, \href{mailto:yufang.hou@it-u.at}{yufang.hou@it-u.at}
 }
}

\begin{document}
\maketitle

\begingroup
\renewcommand\thefootnote{$\dagger$}
\footnotetext{These authors jointly supervised this work.}
\endgroup

\begin{abstract}
Extracting scientific evidence from biomedical studies for clinical research questions (e.g., \emph{Does stem cell transplantation improve quality of life in patients with medically refractory Crohn's disease compared to placebo?}) is a crucial step in synthesising biomedical evidence. In this paper, we focus on the task of document-level scientific evidence extraction for clinical questions with conflicting evidence. 
To support this task, we create a dataset called \textsc{CochraneForest} leveraging forest plots from Cochrane systematic reviews. 
It comprises 202 annotated forest plots, associated clinical research questions, full texts of studies, and study-specific conclusions.
Building on \textsc{CochraneForest}, we propose URCA (Uniform Retrieval Clustered Augmentation), a retrieval-augmented generation framework designed to tackle the unique challenges of evidence extraction. 
Our experiments show that URCA outperforms the best existing methods by up to 10.3\% in F1 score on this task. However, the results also underscore the complexity of \textsc{CochraneForest}, establishing it as a challenging testbed for advancing automated evidence synthesis systems. 
\end{abstract}

\section{Introduction}
Medical practitioners face the challenge of staying current with the ever-growing volume of medical research, making it increasingly difficult to discern meaningful findings from irrelevant ones. Systematic reviews aim to address this challenge by synthesising all relevant evidence for a specific clinical question~\cite{cochranehandbook}, providing clear, up-to-date answers derived from high-quality research. Systematic reviews are widely regarded as the gold standard in evidence-based medicine, heavily influencing medical decisions made by doctors, health authorities, and patients.

However, the process of producing systematic reviews is both time-consuming and costly. A 2019 study estimated that on average conducting a systematic review takes 1--2 years and costs over \$141,000~\cite{michelson2019significant}, due to the extensive effort required to sift through vast amounts of potentially relevant studies and perform rigorous statistical analysis before drawing a final conclusion. Given the substantial resources required, there is growing interest in automating various steps of the 
process~\cite{marshall2019toward, khraisha2023can, yun2023appraising, yun2024automatically}. 

\begin{figure}
    \includegraphics[width=\columnwidth]{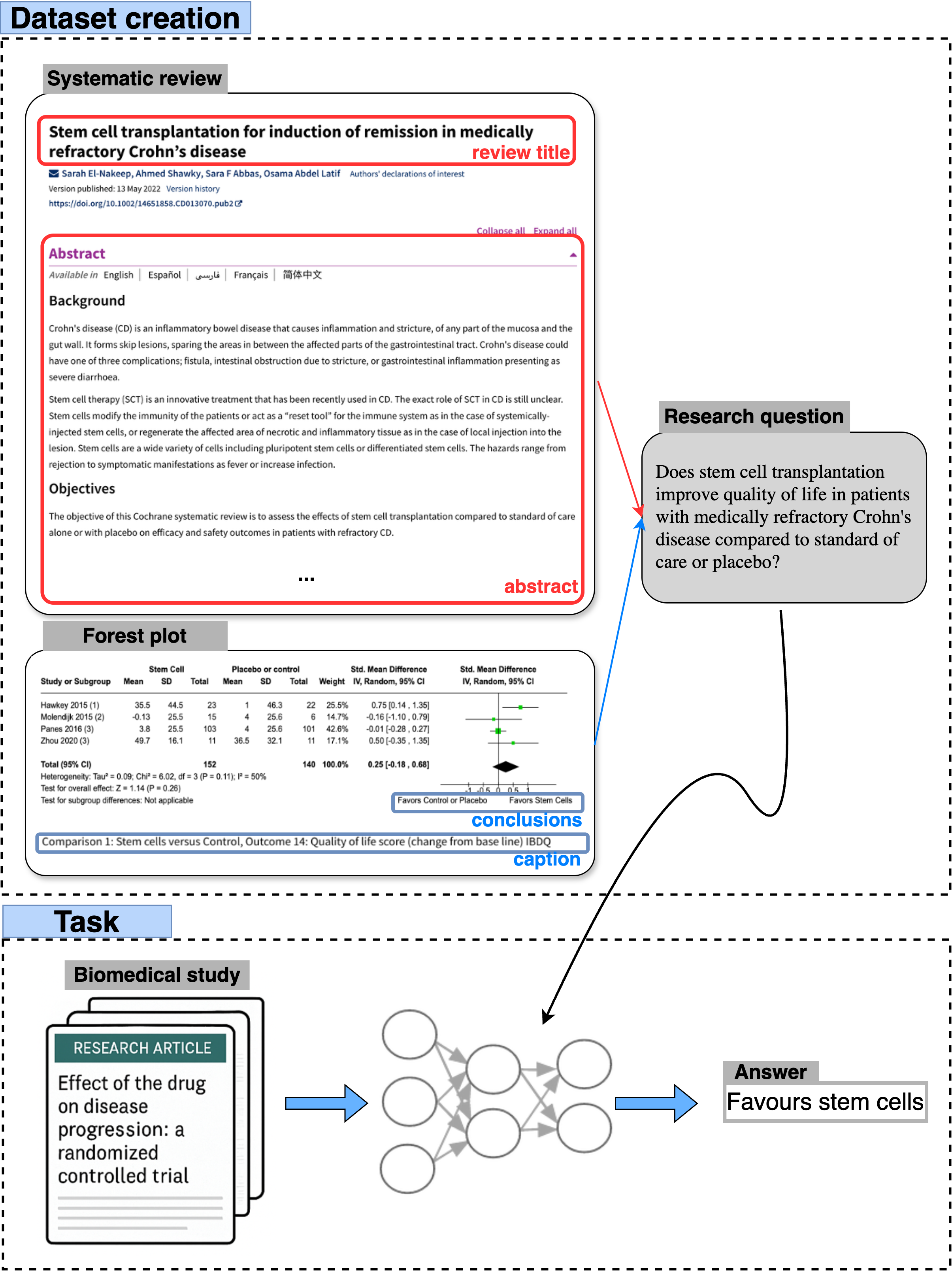}
    \caption{An example from \textsc{CochraneForest}. The research question is annotated based on systematic review context and forest plot, where each row represents
    a study and its conclusion about the question. 
    }
        \label{fig:corpus}
\end{figure}

In this paper, we focus on document-level scientific evidence extraction for clinical questions from biomedical studies that contain contradictory conclusions.
This involves identifying relevant 
information
from the papers comprising a study and deriving the conclusion of the study to the question. To facilitate research on this task, we construct \textsc{CochraneForest}, 
a new dataset containing 202 human-annotated forest plots extracted from 48 real-world systematic reviews and accompanied by the full text of their included 263 unique studies. A forest plot, as illustrated in Figure~\ref{fig:corpus}, is the cornerstone of biomedical systematic reviews, consolidating diverse study results into a single axis. It facilitates direct comparison of findings from different studies and provides a visual representation of statistical analyses. In our work, we leverage forest plots to annotate random control trial (RCT) studies containing contradicting conclusions regarding a research question (Section \ref{sec:dataset}).

To tackle the challenge of extracting scientific evidence from RCTs, which often involve multiple papers reporting results at various stages of the study, 
we propose \textbf{URCA (Uniform Retrieval Clustered Augmentation)}, a retrieval-augmented generation framework designed to address the unique challenges of evidence extraction from individual studies. By uniformly retrieving relevant passages from different papers within the same study and leveraging large language models (LLMs) to distil query-specific information from the clusters of the retrieved passages, URCA addresses common challenges in traditional RAG systems, such as noisy retrieval and the failure to incorporate relevant information from diverse sources (Section \ref{sec:method}).

	
Through extensive experimentation, we demonstrate that URCA outperforms 
strong baselines, 
achieving up to a 
10.3\%
improvement in F1 score for predicting the conclusions of RCT studies for the given questions. 
Our results and analyses demonstrate the importance of the new task and dataset to support significant future research in this domain, as well as the robustness of the proposed method when applied to other question answering (QA) tasks and datasets (Section \ref{sec:exp}).

In summary, our main contributions are: 
(1) we introduce and formalise the task of document-level evidence extraction from RCT studies that present contradictory findings in response to a research question; (2) we propose \textsc{CochraneForest}, a novel dataset derived from forest plots in systematic reviews, encompassing research questions, the corresponding full-text studies, and their conclusions regarding the research question; (3) we develop a novel RAG-based approach, URCA, and establish it as a robust baseline for this task.

\section{Background and Task Definition}
\label{sec:task}
\paragraph{Systematic review.}
A systematic review is a rigorous process for identifying, evaluating, and synthesising evidence from multiple studies to address a specific research question. Biomedical systematic reviews often compare clinical interventions by analysing data from trials or studies to inform 
clinical guidelines.

\paragraph{Papers and studies.} A key distinction exists between \textit{papers} and \textit{studies}. Medical RCT studies often generate substantial data across a long time period, which may be disseminated across multiple publications. During the review process, numerous studies and papers are assessed, but only a subset of studies and their corresponding papers are included in the final synthesis of evidence, referred to as \textit{included studies}.

\paragraph{Forest plots.} 
A forest plot is a graphical representation used in systematic reviews to summarise the results of multiple studies on a particular research question. It visually displays the effect sizes of individual studies along with their confidence intervals, allowing for easy comparison across studies. As shown in Figure~\ref{fig:corpus}, the plot typically includes a central vertical line representing the null effect (e.g., no association), with each study shown as a point estimate and a horizontal line indicating the confidence interval. A summary estimate, often represented by a diamond shape, provides an overall effect size based on all included studies. 


\paragraph{Task definition.}
The inputs to our task are a research question $q$ and a set of studies $\mathcal{S}$ under evaluation in a forest plot $\mathcal{F}$ belonging to a systematic review (Figure~\ref{fig:corpus}). The research question $q$ is formulated to compare the effects of two interventions (e.g., \emph{stem cell transplantation} and \emph{placebo}) for a specific condition (e.g., \emph{patients with medically refractory Crohn's disease}) with regard to a specific outcome measure (e.g., \emph{quality of life}).
Each study $s \in \mathcal{S}$ contains one or more papers $S = \{p_1, ..., p_n\}$ and is associated with a conclusion $c$ with respect to the research question $q$. For each study $s \in \mathcal{S}$, the system must predict the correct conclusion $c$ (e.g., \emph{favours interventions, favours placebo, no difference}) given $q$ and the papers $\{p_1, ..., p_n\}$. 

\section{The \textsc{CochraneForest} Dataset}
\label{sec:dataset}
The \textsc{CochraneForest} dataset consists of 202 forest plots derived from 48 Cochrane\footnote{\url{https://www.cochranelibrary.com}} medical systematic reviews. Each forest plot is characterised by a research question, a set of studies under assessment, and a list of conclusions. In the following sections, we describe the process used to create and annotate this corpus. 


\subsection{Dataset Preparation}\label{ssec:filtering}
To construct \textsc{CochraneForest}, we leveraged the Cochrane Database of Systematic Reviews (CDSR),\footnote{\url{https://www.cochranelibrary.com/cdsr/reviews}} the leading resource for systematic reviews in healthcare containing 9,301 systematic reviews encompassing over 220,000 included studies as of September 2024.

Our construction process was guided by the dual goals of ensuring diversity in the dataset while addressing practical challenges such as accessibility and data quality. To this end, we began by downloading the CDSR database and then applied a series of systematic filtering steps to refine the dataset, as shown 
in Figure ~\ref{fig:filtering-steps} (Appendix \ref{appendix:corpus_construction}).

First, we excluded withdrawn systematic reviews, which comprised approximately 0.5\% of the database, and retained only the latest version of each review to eliminate redundancy caused by prior versions. Reviews with fewer than two included studies were also discarded to ensure that every forest plot in the dataset represented meaningful evidence synthesis. A significant challenge lay in ensuring that all studies referenced in the systematic reviews were accessible in full text. In fact, systematic reviews often include diverse sources such as journal articles, clinical trial reports, and even PhD theses, and only a subset of these are openly available. To address this issue, we retained only reviews for which all included studies were openly accessible. We refer to this subset of reviews as \textit{complete} reviews, which represent the backbone of our corpus. Since we are interested in contradictory scientific evidence, we further filter the corpus to retain the systematic reviews presenting forest plots that contain at least two studies with contradicting conclusions. 
A forest plot is considered to contain contradictory conclusions if at least one study presents findings that differ from those of other studies. For instance, we consider that a study favouring stem cell transplantation contradicts another study reporting no difference between stem cell transplantation and placebo.


After these filtering steps, the final \textsc{CochraneForest} dataset consists of 202 annotated forest plots extracted from 48 systematic reviews, 263 unique studies, and 923 total records (research question-study pairs). It is worth noting that since \textsc{CochraneForest} was built from the whole Cochrane database of systematic reviews, it represents the biggest possible dataset of systematic reviews 
that complies with the open accessibility requirement for the included studies.
Moreover, its scale is comparable to established benchmarks such as BioASQ~\cite{bioasq} (618), PubMedQA~\cite{pubmedqa} (500), and MMLU-Med~\cite{mmmlumed} (1089).
Distribution of studies and papers is shown in Table~\ref{tab:dataset_stats}.

\begin{table}[t]
	\centering
        \scalebox{0.9}{
	\begin{tabular}{lrrr}
		\hline
		& Mean  & Max     & Min  \\ \hline
		Studies per review    &  5.67 & 13 &  2  \\
		Papers per review     &  6.87 & 15 &  2  \\
		Papers per study       &  1.82 &   5 &  1 \\
		Studies per forest plot & 4.57 & 10 & 2 \\ 
		\hline
	\end{tabular}
	}
    \caption{Statistics for the dataset.}
        \label{tab:dataset_stats}
\end{table}

\subsection{Forest Plot Annotation}
For each forest plot, annotators were provided with the corresponding Cochrane systematic review, a corresponding research question, the set of studies included in the forest plot analysis, and the possible conclusions for each study. Each research question was generated by prompting \texttt{llama-3.1-70b} given the title and abstract of the systematic review as well as the caption and the conclusions set of the forest plot. See Appendix \ref{appendix:prompts} for more details.

The annotation tasks were as follows. First, annotators were asked to verify and, when necessary, edit the automatically generated research question to ensure it was consistent with the analysis shown in the forest plot (\textbf{Task 1}, Figure \ref{fig:annot-tasks} in the appendix). This involved ensuring that the question accurately captured the target population, intervention, comparator, and outcome. 

Next, annotators were presented with three predefined labels (i.e., {\emph{favours left intervention, favours right intervention, show no difference between left and right interventions}) for annotating the conclusion of each study, with one label pre-selected based on the automatically extracted 95\% confidence interval (CI) reported in the forest plot (\textbf{Task 2}, Figure \ref{fig:annot-tasks} in the appendix). See Appendix \ref{appendix:ConclusionExtraction} for more details. Notably, no annotator modified the pre-selected label. 

Finally, annotators reviewed and, when necessary, revised the two intervention names
extracted directly from the axes of the forest plot (\textbf{Task 3}, Figure \ref{fig:annot-tasks} in the appendix). We require annotators to rephrase when the original text was deemed insufficiently clear or explicit out of context (e.g., to expand acronyms or clarify ambiguous terms). 
Combining the annotation results from Tasks~2 and~3, we can effectively derive concrete conclusions for each study in a forest plot.

Annotators included two experts with NLP backgrounds, two graduate students studying computer science, and a medical domain expert, all of whom are co-authors of this paper. The annotation interface is shown in Appendix ~\ref{appendix:annotation_interface}.

\subsection{Inter-annotator Agreement}
To assess the reliability of the annotations, 15 forest plots (63 studies in total) were selected and assigned for independent re-annotation by four of the annotators. We exclude Task 2 from the analysis as no annotator modified the pre-selected labels. We computed two kinds of agreement metrics: (1) \textbf{per-annotator metrics}, which evaluate individual annotators' performance relative to the original annotation content; and (2) \textbf{aggregated pairwise metrics}, which assess the consistency between pairs of annotators across all annotation tasks, providing an aggregated measure of agreement. For per-annotator evaluation, we measure the average proportion of items changed $\varphi$, the character-based Levenshtein distance~\cite{levenshtein1966binary} and HTER~\cite{hter} against the original annotation item (Table~\ref{tab:iaa-per-annot} in the appendix). Aggregated agreement is computed as mean pairwise HTER and mean pairwise semantic similarity, where the latter is defined as the cosine similarity between the embedding vectors of the annotation items. Additionally, we report Fleiss' $\kappa$~\cite{fleiss71} on annotators' change labels.

Results (Table~\ref{tab:aggregated-metrics-task}) indicate that, while edit-based metrics yield relatively low scores due to their sensitivity to minor textual changes, the semantic similarity between annotations is notably high, with cosine similarity scores of 0.95 for Task 1 and 0.90 for Task 3. This suggests strong agreement on the texts' underlying meaning. Full metric definitions and additional analyses are provided in Appendix~\ref{appendix:iaa}.  


\begin{table}[t]
	\centering
	\scalebox{0.8}{
	\begin{tabular}{lcc}
		\toprule
		\multirow{2}{*}{\textbf{Metric}} & \multicolumn{2}{c}{\textbf{Score}} \\
		\cmidrule(lr){2-3}
		& \textbf{Task 1} & \textbf{Task 3} \\
		\midrule
		Cosine Similarity ($\overline{s}$) &  0.95 & 0.90 \\
		HTER & 0.34 & 0.36  \\
		Fleiss $\kappa$~\cite{fleiss71}& 0.06 & 0.55 \\
		\bottomrule
	\end{tabular}
	}
	\caption{Aggregated metrics across all annotators for Task 1 and Task 3.}
	\label{tab:aggregated-metrics-task}
\end{table}

\section{URCA: Uniform Retrieval Clustered Augmentation}
\label{sec:method}

In this section, we present URCA (Uniform Retrieval Clustered Augmentation), a novel retrieval-augmented generation (RAG) approach tailored for 
the document-level evidence extraction defined in Section \ref{sec:task}. 

    \begin{figure*}[ht]
        \includegraphics[width=\linewidth, height=.28\textheight]{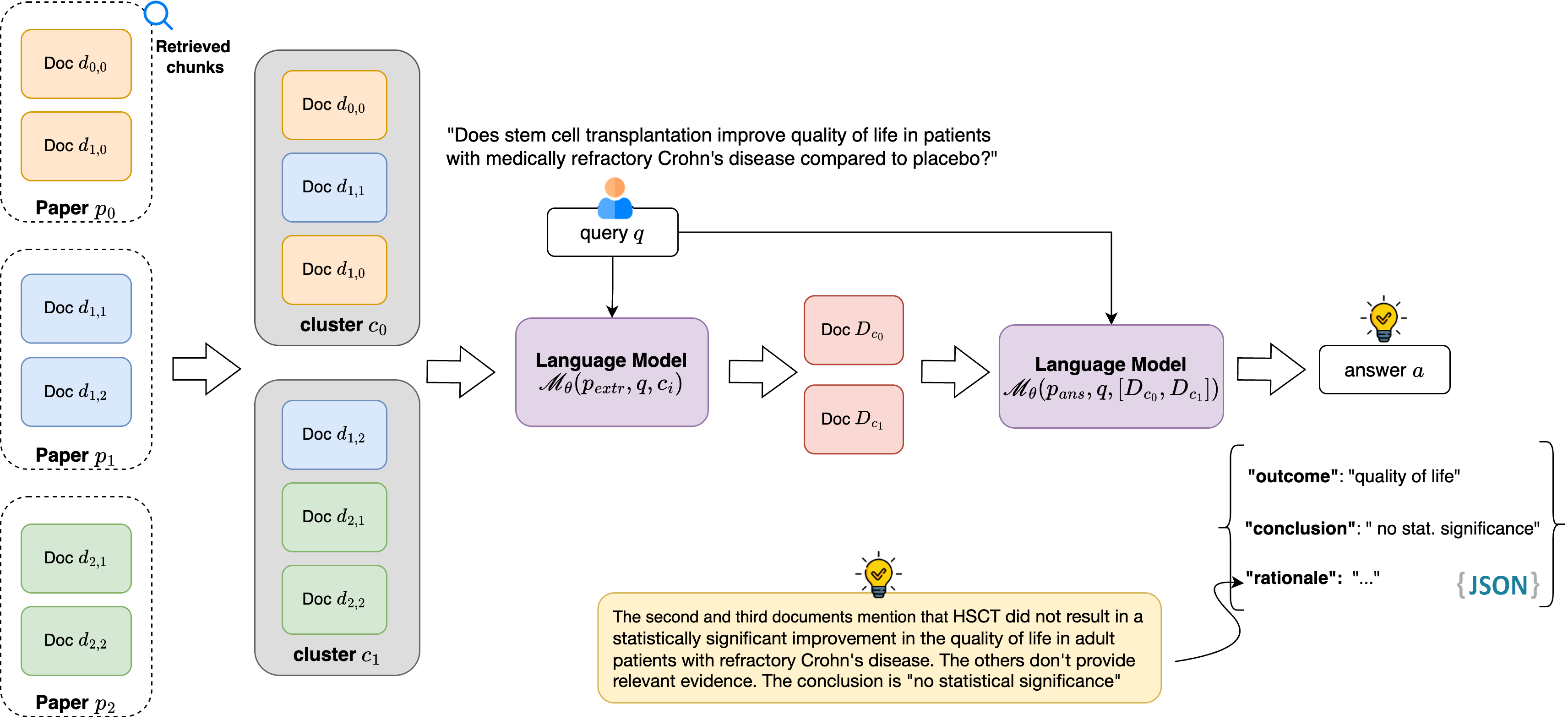} \hfill
        \caption {Overview of the URCA framework. Retrieval size is first distributed across all papers. Retrieved chunks are then clustered and aggregated into passages. Finally, the obtained passages are used to generate the final answer.}
        \label{fig:urca}
    \end{figure*}
    
\subsection{Framework Overview}\label{section:urca-overview}
    As detailed in Algorithm~\ref{alg:urca} and illustrated in Figure~\ref{fig:urca}, URCA starts from uniformly distributing the retrieval size over the sources of interest. Then the retrieved passages are clustered and aggregated by extracting the relevant information to the query using a language model. 
    Eventually, the final answer is generated based on the aggregated information produced in the previous step. 

	\begin{algorithm}
		\caption{URCA}\label{alg:urca}
            \footnotesize
		\begin{algorithmic}[1]
			\Require Query $q$, Desired number of retrieval passages $k$, Scaling factor $\beta$, Sources $S = \{s_1, ..., s_S\}$, Embedding Retriever $\mathcal{R}$, Language model $\mathcal{M}_\theta$, Prompt templates $p_\text{extr}, p_\text{ans}$
			\State $k_s \gets \lceil\min(k + \beta \cdot \log(S), N_{\text{max}})/S\rceil$
			\State $E_T \gets []$ 
			\For{each source $s \in S$}
			\Comment{Uniform retrieval}
			\State $[(e^s_1, t^s_1), ..., (e^s_{k_s}, t^s_{k_s})] \gets \mathcal{R}(q, s_i, k_s)$
			\Comment{Retrieve embeddings and texts}
			\State $E_T \gets E_T \oplus [(e^s_1, t^s_1), ..., (e^s_{k_s}, t^s_{k_s})]$
			\EndFor
			\State $[c_1, ..., c_n] \gets \text{Cluster}(E_T)$
			\Comment{cluster embeddings with UMAP + GMM}
			\For{each cluster $c$}
			\State $D_i \gets \mathcal{M}_\theta(p_\text{extr}, q, c)$
			\Comment{Extract knowledge from clustered texts}
			\EndFor
			\State $a \gets \mathcal{M}_\theta(p_\text{ans}, q, \langle D_1, ..., D_n \rangle)$
			\Comment{Generate the final answer}
			\State \Return $a$
		\end{algorithmic}
	\end{algorithm}

    \subsection{Uniform Retrieval}\label{section:unif-ret}
In the first step, we retrieve the top-$k$ most relevant documents for the query $q$, ensuring balanced representation across the sources of interest (i.e., the study papers). To achieve this, we allocate a portion of $k$ to each source. Specifically, for each source, we retrieve $k_s$ entries as follows:
    \[
        k_s = \lceil \min(k  + \beta \cdot \log(S), N_\text{max}) / S \rceil
    \]
     where $S$ is the number of sources under consideration and $\beta$ is a scaling factor that controls the impact of the logarithmic adjustment. A higher $\beta$ ensures more results are retrieved per source when the number of sources S is large, which is particularly important when $k < S$. Conversely, a lower $\beta$ reduces oversampling, keeping the allocation closer to an even distribution of $k / S$. This approach ensures proportionality while maintaining sufficient representation from each source. We find that this approach is beneficial regardless of the subsequent steps of the pipeline.
    
    \subsection{Clustering and Knowledge Extraction}\label{section:clust}    		
    In the second step, we cluster the embeddings of the retrieved documents and employ an LLM to extract the relevant information from each cluster given the query. Our clustering approach follows the methodology introduced by RAPTOR \cite{raptor}, which employs Gaussian Mixture Models (GMM) for clustering, along with Uniform Manifold Approximation and Projection \cite{umap} for dimension reduction of the dense embeddings, and the Bayesian Information Criterion (BIC) for model selection. 
		
	To aggregate knowledge, we prompt the LLM $\mathcal{M}_\theta$ (with $p_{extr}$ in Appendix~\ref{appendix:prompts}) to identify the relevant evidence for the research question $q$ from each cluster of texts $c_i$. This process discards information irrelevant to $q$,  
    which is particularly valuable for addressing complex questions. The extracted information for each cluster is represented as:
		\[
			D_i = \mathcal{M}_\theta (p_{extr}, q, c_i), \ \ \ i = 1,...,n
		\]
        
    The resulting set of $D_i$ passages can be viewed as distilled, query-aware evidence of each cluster.
    
    \subsection{Answer Finalisation}\label{section:ans}
    In the last step, we prompt the LLM (with prompt $p_{ans}$ in Appendix~\ref{appendix:prompts}) to generate the final answer given the research question $q$ and the passages $\langle D_1, ..., D_n \rangle$ produced at the previous step:
    \[
        a = \mathcal{M}_\theta (p_{ans}, q, \langle D_1, ..., D_n\rangle)
    \]

    This step combines the distilled insights from each cluster into a coherent and concise answer. By deferring final synthesis until after clustering and extraction, we ensure that the model operates on already-filtered and semantically aligned information, reducing noise and improving accuracy.

\section{Experiments}
\label{sec:exp}
\subsection{Experimental Setup}
\paragraph{Problem formulation.} 
We adopt the standard RAG setting where the LLM $M_\theta$ has access to an external knowledge base through an off-the-shelf retriever $\mathcal{R}$. Differently from existing approaches, we assume to have pre-filtered the sources $S$ relevant to a research question $q$. These are the papers composing the studies under assessment. Given $q, \mathcal{R}$ and a set of possible conclusions $\mathcal{C}$, the goal is to predict the correct conclusion $c \in \mathcal{C}$ to $q$. 

Notably, we directly employ off-the-shelf retrievers instead of training our own, and prepend all retrieved documents to the question as input to the model, without any re-ranking. This setting is orthogonal to existing research efforts centered on improving the retriever or performing adaptive retrieval~\cite{rat, selfrag}.

\paragraph{Baselines.} To evaluate URCA on \textsc{CochraneForest}, we compare it against a diverse set of baselines testing different retrieval and synthesis strategies. As a starting point, we include two baselines that eliminate active retrieval: (1) No RAG which relies solely on the model's internal knowledge, and (2) Abstracts which injects study abstracts directly into the context. These test the importance of external evidence acquisition.
Moreover, we include vanilla RAG, a straightforward retrieval-augmented generation approach, which we evaluate both with and without uniform retrieval to highlight the limitations of simpler methods and the potential benefits of incorporating a more balanced retrieval strategy. Additionally, we include \textsc{InstructRAG}~\cite{instructrag}, a more advanced variant that prompts the model to provide rationales connecting answers to the retrieved evidence in passages. For a fair comparison, we use the instructions without training or in-context learning. RAPTOR~\cite{raptor}, the state-of-the-art method that uses recursive clustering and summarisation to synthesise information, serves as a robust comparison for tasks requiring multi-source evidence synthesis. Lastly, we include \textsc{GraphRAG}~\cite{graphrag}, which builds a graph-based text index by summarising closely 
related entities 
from the source documents.

\paragraph{Other datasets.} While URCA is designed to leverage the structured nature of systematic reviews, its underlying approach can be applicable to broader question-answering tasks. To explore this possibility, we conduct experiments on two widely used datasets: 
PubMedQA~\cite{pubmedqa} and MedQA-US~\cite{medqa}. In these evaluations, we removed the uniform retrieval step for two key reasons. First, uniform retrieval is most beneficial when the relevant sources are known in advance, as is the case for systematic reviews, but this assumption does not hold for open-domain QA tasks. Second, clustering provides greater performance gains than uniform retrieval (Table~\ref{tab:ablation-1}). 

\begin{table*}[h]
	\small
        \centering
	\renewcommand{\arraystretch}{1.1} 
	\aboverulesep = 0pt
	\belowrulesep = 0pt
	\begin{tabular}{lcccc|cccc}  
		\toprule
				\addlinespace[0.5ex] 
		\textbf{Method}  & F1 & Precision & Recall & Accuracy  & F1 & Precision & Recall & Accuracy \\   
		\midrule
		\addlinespace[0.5ex] 
		& \multicolumn{4}{c|}{\cellcolor{gray!20} \textbf{Mistral-Large (2407), 128B}} & \multicolumn{4}{c}{\cellcolor{gray!20} \textbf{LlaMa 3.1, 70B}} \\
		\addlinespace[0.5ex] 
		\midrule
		No RAG & 46.1 & 47.4 & 44.8 & 53.1 & 49.1 & 46.1 & 52.4 & 45.2  \\
		Abstracts & 62.6 & 60.3 & 65.0 & 62.6 & 60.7 & 60.4 & 61.1 & 64.1  \\
		\midrule
		RAG  & 60.9 & 60.7 & 61.0 & 64.3 & 62.1 & 60.2 & 64.1 & 62.9  \\
		 \hspace{.9cm}+ Uniform Retrieval & 63.3 & 60.9 & 65.8 & 63.2 & 63.4 & 61.5 & 65.5 & 64.2  \\
		 \hline
		RAPTOR  & 61.7 & 59.1 & 64.6 & 61.0 & 60.6 & 58.2 & 63.2 & 60.0  \\
		InstructRAG  & 62.6 & 61.5 & 63.6 & 64.4 & 60.9 & 59.6 & 62.3 & 62.1  \\
		GraphRAG  & 64.9 & 63.8 & 66.0 & 66.3  & 65.6 & \textbf{64.8} & 66.3 &  63.4 \\
		\midrule
		\textbf{URCA (Ours)} & \textbf{67.3} & \textbf{65.2} & \textbf{69.5} & \textbf{67.0} & \textbf{66.1} & 64.0 & \textbf{68.4} & \textbf{64.6} \\
		\midrule
		\addlinespace[0.5ex] 
		& \multicolumn{4}{c|}{\cellcolor{gray!20} \textbf{GPT-4 (0613)}} & \multicolumn{4}{c}{\cellcolor{gray!20} \textbf{GPT-3.5-Turbo (0613)}}\\  
		\addlinespace[0.5ex] 
		\midrule
		No RAG & 47.5 & 55.9 & 41.2 & 59.5 & 24.1 & 18.9 & 33.2 & 56.4  \\
		Abstracts & 61.0 & 60.2 & 61.9 & 62.6 & 56.0 & 55.2 & 56.9 & 58.8  \\
		\midrule
		RAG  & 61.6 & 59.3 & 64.0 & 61.6 & 59.1 & 57.6 & 60.6 & 60.7 \\
		\hspace{.9cm}+ Uniform Retrieval & 62.0 & 59.7 & 64.5 & 61.8 & 61.8 & 60.0 & 63.8 & 62.8  \\
		\hline
		RAPTOR  & 60.1 & 58.2 & 62.0 & 60.8 & 53.6 & 52.0 & 55.3 & 54.1  \\
		InstructRAG  & 61.6 & 61.4 & 61.9 & 63.3 & 57.4 & 57.1 & 57.7 & 61.1  \\
		GraphRAG  & 63.8 & 61.6  & 66.1 & 63.7 & 56.6  & 58.4  &  54.9 & 59.3  \\
		\midrule
		\textbf{URCA (Ours)} & \textbf{65.7} & \textbf{63.0} & \textbf{68.7} & \textbf{64.1} & \textbf{62.4} & \textbf{60.6} & \textbf{64.4} & \textbf{63.5} \\
		\bottomrule
	\end{tabular}
\caption{Overall results of URCA and seven baselines on \textsc{CochraneForest} on four LLMs, showing micro-F1 score, micro-precision, micro-recall, and accuracy. Best scores are reported in bold. }
\label{tab:main-results}
\end{table*}

\begin{table}
\centering
\scalebox{0.80}{
\begin{tabular}{lcc}
    \toprule
    \textbf{Method} & \textbf{MedQA-US} & \textbf{PubMedQA} \\
    \midrule
    No RAG & 72.1 & 77.5 \\
    RAG & 82.3	& 79.6  \\
    GraphRAG & 84.5 & 80.6  \\
    URCA & \textbf{85.9} & \textbf{81.1} \\
    \bottomrule
\end{tabular}
}
\caption{Accuracy on MedQA-US and PubMedQA.}
\label{tab:other-med-datasets}
\end{table}

\paragraph{LLM and RAG settings.}\label{rag-settings} We conduct experiments on both open and closed-source LLMs of different sizes, including \texttt{Llama-3.1-70B}, Mistral Large (\texttt{Mistral-Large-Instruct-2407}), GPT 3.5 Turbo (\texttt{gpt-3.5-turbo-0613}), and GPT 4 (\texttt{gpt-4-0613}). The generation temperature is set to 0, and the maximum output tokens is set to 1,024. By default, we use the top 10 retrieved passages in all the approaches under comparison.

\subsection{Main Results}

\paragraph{Performance on \textsc{CochraneForest}.} Table~\ref{tab:main-results} presents the results on \textsc{CochraneForest} for each model and approach under evaluation. Comparing No RAG and the various retrieval-based methods, we observe that relying solely on the model’s internal knowledge is insufficient for this task, highlighting the necessity of external evidence to address domain-specific questions. In contrast, the Abstracts approach emerges as a surprisingly strong baseline. The improvements achieved by standard RAG or even more sophisticated approaches like RAPTOR and \textsc{InstructRAG} are relatively small in comparison, underscoring the value of abstracts as a compact and effective evidence source. However, abstracts alone are ultimately not enough to address the level of granularity required for outcome-specific questions, which often rely on access to tables, analyses, and other fine-grained evidence beyond the abstract. 

Interestingly, RAG and RAG + Uniform Retrieval demonstrate clear improvements over the previous baselines and over more sophisticated approaches such as \textsc{InstructRAG}. This can be attributed to the latter’s reliance on carefully curated demonstrations, a limitation also noted by~\citet{astuteRAG} for domain-specific datasets like BioASQ~\cite{bioasq}.
RAPTOR, despite sharing conceptual similarities with URCA, also performs poorly overall. This behaviour can be explained by its offline clustering and synthesis processes which are not guided by the query. As a result, it risks focusing on irrelevant outcome measures or introducing information loss, thus misleading the LLM, ultimately limiting its effectiveness. Similar considerations can be made for  \textsc{GraphRAG}, which relies on multiple steps of entity linking and offline summarisation. 

This underscores a critical insight from our findings: more sophisticated RAG variants are not inherently better in domain-specific biomedical tasks unless their design is explicitly aligned with the evidence structure of the data.

Overall, URCA delivers the strongest performance across all models and metrics on \textsc{CochraneForest}. In particular, URCA improves the best baseline (GraphRAG) by at least 3\%  relative F1 on Mistral Large and GPT-4, reaching a peak of 10.3\% on \texttt{gpt-3.5-turbo}.

\begin{tiny}
\begin{table*}[httb]
	\centering
	\scalebox{0.8}{%
		\begin{tabular}{llSSSS}
			\toprule
			\textbf{Model}         & \textbf{Method}                 & \textbf{F1} & \textbf{Precision} & \textbf{Recall} & \textbf{Accuracy} \\ 
			\midrule
			
			\multirow{3}{*}{\textbf{Llama-3.1-70B}} 
			& URCA                           &  66.1 & 64.0 & 68.4 & 64.6    \\
			& \hspace{1cm}w/o Uniform Retrieval &   64.5~\textcolor{red2}{\text{($\downarrow$ 2.4\%)}}      & 63.3 & 65.9 & 64.6        \\
			& \hspace{1cm}w/o Clustering &   63.2~\textcolor{red2}{\text{($\downarrow$ 4.5\%)}}      & 61.9 & 64.5 & 63.6      \\
			\midrule
			
			\multirow{3}{*}{\textbf{Mistral-Large-2407}} 
			& URCA                           & 67.3 & 65.2 & 69.5 & 67.0       \\
			& \hspace{1cm}w/o Uniform Retrieval & 66.17~\textcolor{red2}{\text{($\downarrow$ 1.6\%)}}        & 64.4 & 68.1 & 66.5
			        \\
			& \hspace{1cm}w/o Clustering &   64.6~\textcolor{red2}{\text{($\downarrow$ 3.9\%)}}      & 63.0 & 66.4 & 64.9        \\
			\midrule
			
			
			\multirow{3}{*}{\textbf{GPT-3.5-turbo-0613}} 
			& URCA                          & 62.4 & 60.6 & 64.4 & 63.5       \\
			& \hspace{1cm}w/o Uniform Retrieval &    61.0~\textcolor{red2}{\text{($\downarrow$ 2.2\%)}}      & 61.5 & 60.5 & 62.6        \\
			& \hspace{1cm}w/o Clustering &  59.4~\textcolor{red2}{\text{($\downarrow$ 4.8\%)}}    & 59.1 & 59.8 & 61.2      \\
			\midrule
			
			\multirow{3}{*}{\textbf{GPT-4-0613}} 
			& URCA                           & 65.7 & 63.0 & 68.7 & 64.1        \\
			& \hspace{1cm}w/o Uniform Retrieval &   64.7~\textcolor{red2}{\text{($\downarrow$ 1.6\%)}}    & 62.1 & 67.5 & 63.2     \\
			& \hspace{1cm}w/o Clustering &  61.9~\textcolor{red2}{\text{($\downarrow$ 5.7\%)}}       & 59.4 & 64.7 & 60.3
			       \\
			\bottomrule
		\end{tabular}%
	}
	\caption{Ablation studies on the impact of uniform retrieval and clustering on URCA.}
	\label{tab:ablation-1}
\end{table*}
\end{tiny}

\paragraph{Performance on open-domain medical QA.} Table~\ref{tab:other-med-datasets} presents the results obtained using GPT-4 under the same experimental settings on two popular medical QA datasets: MedQA-US and PubMedQA. We observe that URCA outperforms the baselines under comparison, demonstrating the robustness and generalisability of our approach beyond the domain of systematic reviews. These results demonstrate that while URCA is tailored for structured biomedical evidence synthesis, its underlying principles can generalise well to broader biomedical QA tasks. On PubMedQA, which focuses on yes/no/maybe questions grounded in abstracts, URCA achieves 81.1\% accuracy, surpassing both GraphRAG and standard RAG by a notable margin. This suggests that even in cases where the input context is less structured than full-text reviews, the clustering mechanism still facilitates better information aggregation and reasoning.

The gains are even more pronounced on MedQA-US, a challenging dataset based on US medical licensing exam questions. These questions are longer, more nuanced, and require multi-hop reasoning across biomedical knowledge. URCA achieves 85.9\% accuracy, outperforming GraphRAG by 1.4\% and standard RAG by 3.6\%.

We attribute URCA’s strong cross-domain performance to its design, which inherently filters and restructures the retrieved content, disentangling overlapping concepts and offering a more interpretable intermediate representation to the model.

\subsection{Ablation Studies and Analyses}\label{ablations}

\paragraph{Influence of components.} As shown in Table~\ref{tab:ablation-1}, we ablate URCA's design from two aspects: (1) w/o uniform retrieval, where the retriever only fetches the top chunks from the knowledge base regardless of the source; (2) w/o clustering, where the knowledge extraction step is performed without clustering the retrieved passages. We observe that both uniform retrieval and clustering play important roles in URCA's design, with clustering having a more pronounced impact. Removing clustering consistently leads to larger performance drops ($4-6\%$ in F1), underscoring its critical role in organising retrieved content for effective knowledge extraction. In contrast, the absence of uniform retrieval results in smaller but noticeable declines ($1.0-2.5\%$), highlighting its role in ensuring diverse coverage. These results confirm that clustering is the primary driver of URCA's performance, while uniform retrieval further enhances its robustness. Figure~\ref{fig:coverage} further supports this by showing that uniform retrieval improves the coverage rate of multiple sources, mitigating biases where one or two papers dominate the retrieved context. This is particularly important in our setting, where individual studies may be documented by a varying number of papers.

\begin{figure}
    \centering
    \includegraphics[width=0.7\linewidth]{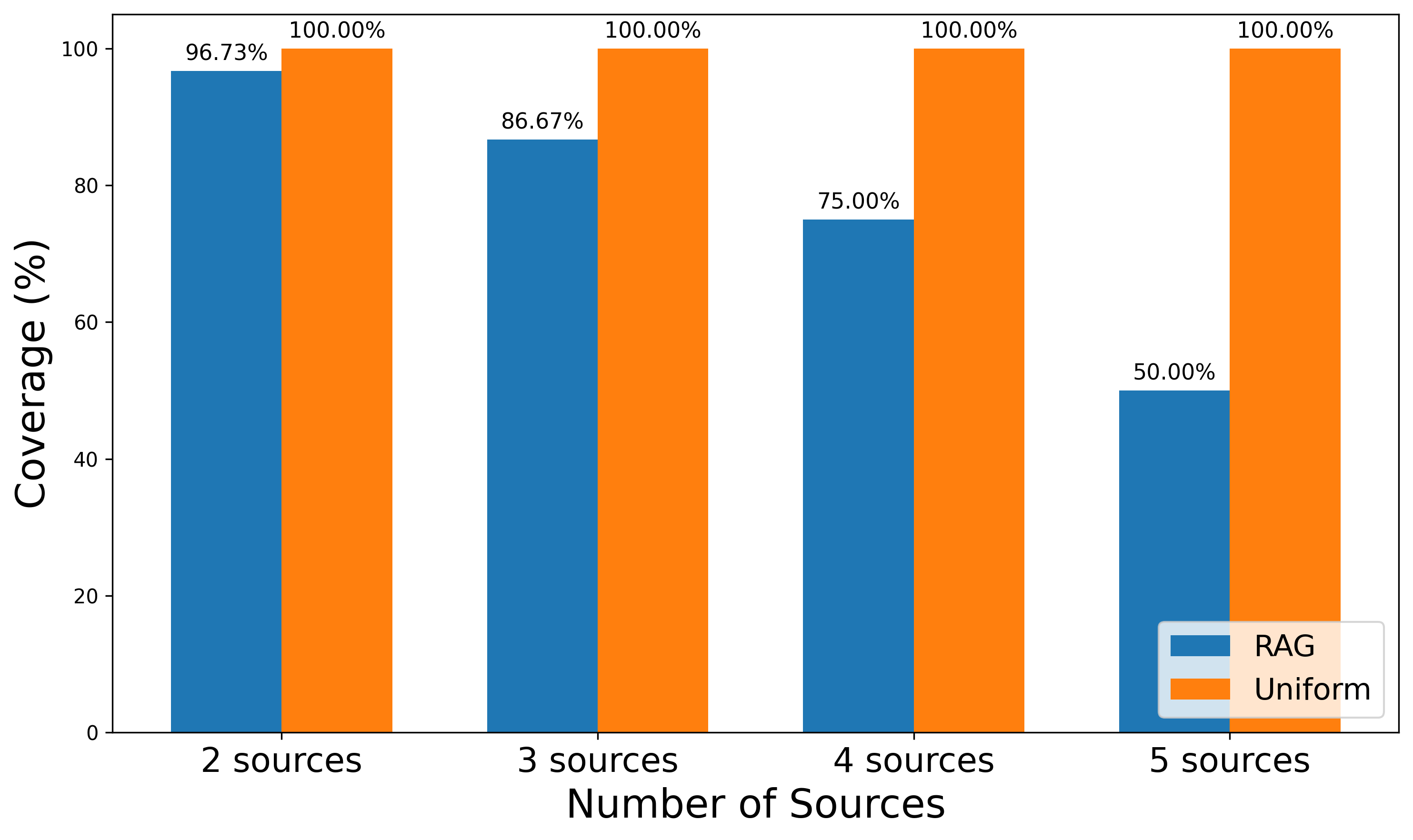}
    \caption{Coverage rate of multiple sources with and without uniform retrieval.}
    \label{fig:coverage}
\end{figure}

\paragraph{Influence of cluster ordering.} Following the passage ordering strategies defined by \citet{alessio24}, we assess the impact of cluster reordering in URCA. In particular, we compare five strategies: ascending or descending similarity to the query, random ordering, and two ping-pong variants. The obtained results are reported in Table~\ref{tab:ablation-2}. We observe that the performance improvement of URCA compared to baseline methods is significantly greater than the performance variations observed due to changes in cluster order within URCA. 
In addition, thanks to the knowledge extraction step, URCA proves to be robust to the chosen ordering approach, with only minor performance variations of $\pm1$\%. 


\begin{table*}
	\centering
	\scalebox{0.8}{
		\begin{tabular}{llcccc}
			\toprule
			\textbf{Model}         & \textbf{Ordering Strategy}                 & \textbf{F1} & \textbf{Precision} & \textbf{Recall} & \textbf{Accuracy} \\ 
			\midrule
			\multirow{5}{*}{\textbf{Llama-3.1-70B}} 
			& Ascending        &  66.1 & 63.4 & 68.4 & 64.6    \\
			& Descending      & 67.1 & 64.7 & 69.6 & 66.0        \\
			& Random            & 66.1 & 63.9 & 68.6 & 65.2        \\
			& Ping-pong Descending Top-to-bottom & 66.0 & 63.7 & 68.5 & 64.6        \\
			& Ping-pong Descending Bottom-to-top  & 66.8 & 64.5 & 69.4 & 65.7        \\
			\midrule
			\multirow{5}{*}{\textbf{Mistral-Large-2407}} 
			& Ascending   & 67.3 & 65.2 & 69.5 & 67.0      \\
			& Descending & 68.1 & 65.7 & 70.7 & 67.2        \\
			& Random  & 67.6 & 65.1 & 70.2 & 66.6        \\
			& Ping-pong Descending Top-to-bottom & 67.5 & 65.1 & 70.1 & 66.5        \\
			& Ping-pong Descending Bottom-to-top  & 67.5 & 65.1 & 70.2 & 66.5        \\
			\midrule
			
			
			\multirow{5}{*}{\textbf{GPT-3.5-turbo-0613}} 
			& Ascending      & 62.4 & 60.6 & 64.4 & 63.5    \\
			& Descending    & 62.5 & 61.2 & 64.0 & 63.6      \\
			& Random  & 61.2 & 59.8 & 62.7 & 61.8      \\
			& Ping-pong Descending Top-to-bottom & 61.9 & 60.4 & 63.5 & 62.3        \\
			& Ping-pong Descending Bottom-to-top  & 61.9 & 60.4 & 63.5 & 62.3       \\
			\midrule
			
			\multirow{5}{*}{\textbf{GPT-4-0613}} 
			& Ascending                           & 65.7 & 63.0 & 68.7 & 64.1    \\
			& Descending & 64.7 & 61.7 & 67.9 & 62.4        \\
			& Random  & 64.4 & 61.5 & 67.5 & 62.0        \\
			& Ping-pong Descending Top-to-bottom & 64.5 & 61.6 & 67.8 & 62.0       \\
			& Ping-pong Descending Bottom-to-top  & 64.6 & 61.7 & 67.8 & 62.0        \\
			\bottomrule
		\end{tabular}
	}
    \caption{Ablation study on cluster ordering strategies for URCA. ``Ascending" is used in our main experiments.}
        \label{tab:ablation-2}
\end{table*}

\paragraph{Benefits of clustering.} We perform a quantitative analysis in which we replace clustering with contiguous grouping, i.e. after randomly shuffling the retrieved chunks, we naively group contiguous chunks. Figure~\ref{fig:contig-clust} shows that grouping the chunks without a criterion worsens the overall performance.  
This highlights the importance of structured grouping in preserving information quality, as naive contiguous grouping disrupts coherence making it harder for the model to identify consistent signals, whereas clustering helps identify meaningful relationships between chunks and better model judgment in the information extraction step.

%
%
\paragraph{Qualitative Example.} Figure~\ref{fig:qualitative-example} presents a qualitative example from \textsc{CochraneForest} in which we show the intermediate steps of URCA and how the clustering steps improve the final prediction. Without clustering, the LLM doesn't properly handle the context and extracts information regarding unrelated outcomes, thus reaching a wrong conclusion. In contrast, by grouping and filtering chunks within clusters, URCA provides better judgment of ambiguous or mixed data and discards irrelevant information early. Further details about this example can be found in Appendix~\ref{appendix:example}.

\begin{figure}
    \centering
\includegraphics[width=.95\columnwidth]{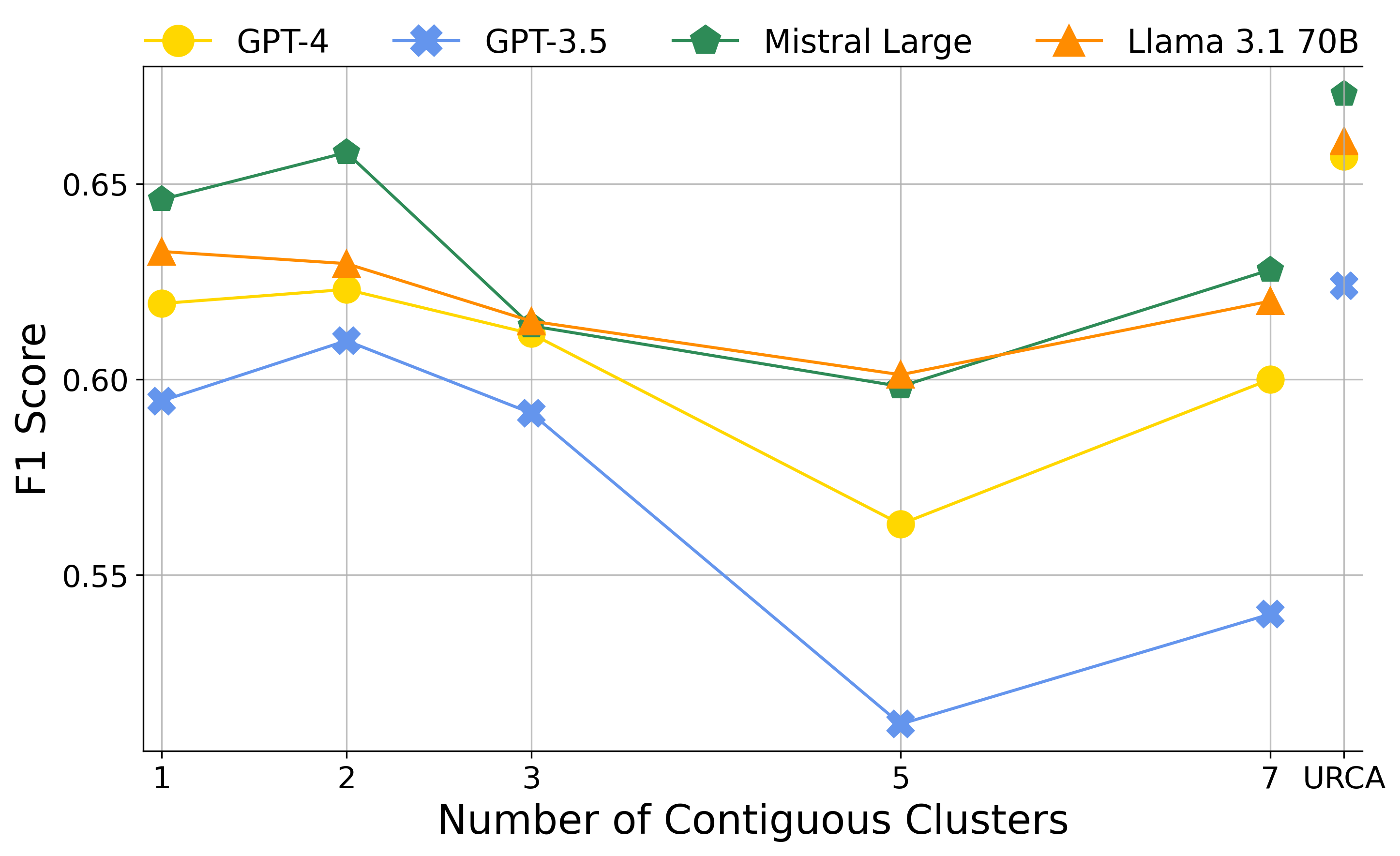}
    \caption{Performance with contiguous grouping versus URCA on different numbers of contiguous clusters.}
    \label{fig:contig-clust}
\end{figure}

\section{Related work}
\paragraph{Medical evidence extraction.}
Previous studies on medical evidence extraction mainly focus on the abstract or individual passage level, such as inferring the effectiveness of a treatment for a given condition \cite{nye-etal-2020-trialstreamer}, extracting contradictory claims about COVID-19 drug efficacy \cite{sosa-etal-2023-detecting}, and summarising single or multiple RCT studies \cite{shaib-etal-2023-summarizing, odoherty-etal-2024-beyond}. Similar to our task, \newcite{lehman-etal-2019-inferring} propose a task to infer reported findings from a
full-text RCT article given a research question. 
In contrast, our work can be seen as an example of scientific argument mining at the global discourse level \cite{al-khatib-etal-2021-argument}. It focuses on extracting medical evidence for a given question across multiple full-text documents, emphasising cross-document synthesis at the full-document level. 
Additionally, our research questions are naturally derived from systematic reviews, aligning closely with real-world clinical research interests.

\paragraph{RAG.} Existing RAG methods~\cite{rag0, rag1, rag2}  have shown promise in improving the factual accuracy of large language models (LLMs) by leveraging up-to-date information and specialised (non-parametrised) knowledge from external sources~\cite{vu2023freshllms, kasai2024realtime}. However, these methods face significant challenges due to the inclusion of irrelevant or erroneous content from retrieval systems~\cite{khattab2022demonstrate, chen2024benchmarking, robustrag} and the inherent noise in retrieval corpora~\cite{izacard2022unsupervised, shao2024scaling, dai2024unifying}. Critically, most existing approaches do not take into account the importance of pre-selecting and diversifying sources. 
	URCA addresses these limitations by 
ensuring all relevant sources are represented at the retrieval stage. 

\section{Conclusion}
We formalised the complex and challenging task of document-level evidence extraction for clinical questions,
also releasing a dataset (\textsc{CochraneForest}) to support further work on this task. We also proposed URCA, a novel RAG-based framework that retrieves uniformly from multiple sources, clusters relevant evidence, and generates conclusions. Our empirical results show that URCA consistently outperforms state-of-the-art RAG approaches 
though performance still leaves room for improvement, highlighting the task's complexity.
Future work could take a more quantitative approach, extracting numerical data to compute 95\% confidence intervals for final conclusions.

\section{Limitations}
Despite the contributions of this work, several limitations must be acknowledged and addressed in future research. First, in the proposed \textsc{CochraneForest} dataset, we annotate research questions and study conclusions but do not provide rationales identifying which specific passages within the included studies support these conclusions. This omission limits the interpretability of the annotations and prevents a fine-grained understanding of the reasoning underlying the conclusions. The absence of rationale annotations also limited our ability to conduct a thorough error analysis. Without identifying supporting evidence, we could not examine where and why models fail when extracting conclusions.


\bibliography{custom}

\appendix

\section{Dataset Construction and Annotation Protocols}
\label{appendix:dataset-annotation}

\subsection{Filtering Systematic Reviews}
\label{appendix:corpus_construction}

\begin{figure}[h]
	\centering
	\includegraphics[width=0.8\columnwidth]{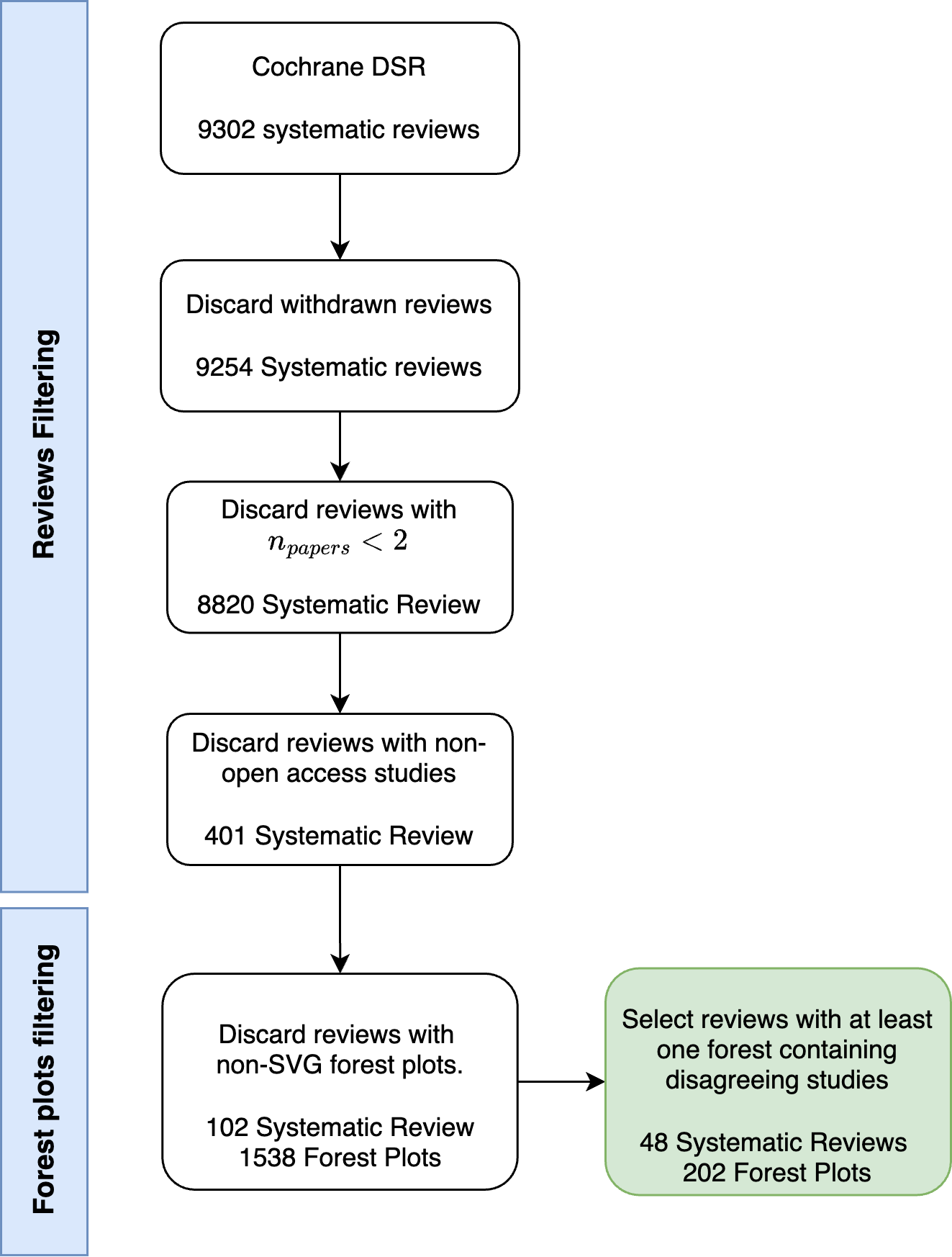}
	\caption{Filtering steps.}
	\label{fig:filtering-steps}
\end{figure}

Figure~\ref{fig:filtering-steps} illustrates the filtering process used to construct the \textsc{CochraneForest} dataset. After downloading the full Cochrane CDSR archive, we applied multiple filtering stages described in Section~\ref{ssec:filtering} to ensure that only reviews with complete data and high-quality forest plots were retained.

\subsection{Inter-Annotator Agreement}
\label{appendix:iaa}

\paragraph{Per-annotator metrics.}  
For each annotator $a_i$ and set of annotation tasks $\mathcal{T}$, we compute the average proportion of items changed $\varphi$, the character-based Levenshtein distance~\cite{levenshtein1966binary} and HTER~\cite{hter} against the original annotation item.  Specifically, for an individual annotator $a_i$, let $\mathcal{T}$  denote the set of annotation tasks (e.g. $\mathcal{T}_1$ for \textbf{Task~1}, $\mathcal{T}_3$ for \textbf{Task~3}) and $\mu(a_i)$ be one of such metrics. Then $\mu(a_i)$ is defined as the average over all annotation items $t_k \in \mathcal{T}$:
\[
	\mu(a_i) = \frac{1}{|\mathcal{T}|} \sum_{t_k \in \mathcal{T}} f(a_i, t_k)
\]   

where $f(a_i, t_k)$ is the function specific to the metric being evaluated. For instance, $f(a_i, t_k) = 1$ if annotator $a_i$ changed the item $t_k$, 0 otherwise. Table~\ref{tab:iaa-per-annot} presents the per-annotator results.

\paragraph{Aggregated pairwise metrics.} Let $\{a_1, ..., a_m\}$ be the set of annotators who annotated the annotation tasks in $\mathcal{T}$. For each pair of annotators $(a_i, a_j)$ and metric $f (a_i, a_j, t_k)$, we define the pairwise agreement as the average of $f$ across all annotation tasks $t_k \in \mathcal{T}$:
\[
	\mu(a_i, a_j) = \frac{1}{|\mathcal{T}|} \sum_{t_k \in \mathcal{T}} f(a_i, a_j, t_k)
\]
The final aggregated agreement for a task set $\mathcal{T}$ is computed as the mean across all annotator pairs:
\[
	\overline{\mu}(\mathcal{T}) = \frac{1}{{m \choose 2}} \sum_{1 \leq i \leq j \leq m} \mu(a_i, a_j)
\]
For this evaluation, we compute again HTER and the semantic similarity expressed as cosine similarity of the embedding vectors $e^i_{t_k}, e^j_{t_k}$:
\vspace{-.3cm}
\[
s(a_i, a_j) = \frac{1}{|\mathcal{T}|} \sum_{t_k \in \mathcal{T}} \frac{e^i_{t_k} \cdot e^j_{t_k}}{\lVert e^i_{t_k} \rVert \lVert e^j_{t_k} \rVert}
\]
Each pairwise similarity $s(a_i, s_j)$ is computed averaging the similarities obtained from 3 embedding models of different sizes selected from the MTEB benchmark~\cite{mteb}: \texttt{all-MiniLM-L12-v2}~\cite{sentence-transformers}, \texttt{NV-Embed-v2}~\cite{nvembed}, and \texttt{text-embeddings-ada-002}~\cite{ada002}. 

Average pairwise metrics are reported in Table~\ref{tab:pairwise-cos-sim} (cosine similarity), Table~\ref{tab:pairwise-frac} (fraction of changes $\varphi$), Table~\ref{tab:pairwise-lev} (Levenshtein distance) and Table~\ref{tab:pairwise-hter} (HTER).

\begin{table}
\centering
	\resizebox{\columnwidth}{!}{
	\begin{tabular}{llccc}
		\toprule
		\textbf{Annotator} & \textbf{Task} & \textbf{$\varphi$} & \textbf{Levenshtein} & \textbf{HTER} \\
		\midrule
		\multirow{2}{*}{A} & Task 1 & 0.33 & 17.60 & 0.10 \\
		& Task 3 & 0.07 & 7.73 & 0.00 \\
		\midrule
		\multirow{2}{*}{B} & Task 1 & 0.87 & 41.93 & 0.49 \\
		& Task 3 & 0.40 & 13.20 & 0.62 \\
		\midrule
		\multirow{2}{*}{C} & Task 1 & 1.00 & 50.73 & 0.50 \\
		& Task 3 & 0.40 & 13.27 & 0.70 \\
		\midrule
		\multirow{2}{*}{D} & Task 1 & 0.87 & 33.07 & 0.34 \\
		& Task 3 & 0.47 & 14.93 & 0.66 \\
		\bottomrule
	\end{tabular}
	}
	\caption{Per-annotator scores for Task 1 and Task 3 for the original text; $\varphi$ is the average proportion of items changed.}
	\label{tab:iaa-per-annot}
\end{table}

\subsection{Annotation Examples}
\label{appendix:example}

\paragraph{Research question annotation.} For research question refinement, annotators were asked to clarify and expand automatically generated questions. For instance, an initial question such as “Does FMT increase the resolution of recurrent \textit{Clostridioides difficile} infection in immunocompetent individuals?” was rewritten expanding acronyms and adding the missing comparator as “Does fecal microbiota transplantation (FMT) increase the resolution of recurrent \textit{Clostridioides difficile} infection in immunocompetent individuals compared to placebo or no treatment?” (Figure~\ref{fig:annot-example}).

\paragraph{Study label annotation.} In labeling studies, annotators examined each forest plot and determined whether individual studies supported the intervention, the control, or remained inconclusive. These decisions were made by visually assessing whether the 95\% confidence intervals crossed the effect threshold indicated in the plot.

\paragraph{Conclusion annotation.} For conclusion annotation, annotators revised and expanded the textual conclusions provided by the original review authors. In cases where the conclusions included acronyms or vague references, these were clarified. For example, “Favours FMT” was revised to “Favours fecal microbiota transplantation (FMT)”, and “Favours control” became “Favours control group” (Figure~\ref{fig:annot-example}).

\section{Extracting Study-Level Conclusions from Forest Plots}
\label{appendix:ConclusionExtraction}

In this section, we describe the method used to determine the overall conclusion of each study in a forest plot, specifically whether the study supports the \textit{intervention}, \textit{control}, or indicates no significant difference, also known as the \textit{null hypothesis}. This process is based on extracting and interpreting the 95\% confidence interval (CI) for each study, which is typically represented as a point estimate (e.g., mean difference, odds ratio, or risk ratio) accompanied by a horizontal line indicating the 95\% confidence interval. The threshold value plays a crucial role in this analysis; it is generally set at \textbf{1} for ratios (e.g., odds ratios, risk ratios) and \textbf{0} for mean differences, representing the null hypothesis, where there is no difference between the intervention and control groups.

To determine the conclusion for each study in a forest plot, we followed a systematic approach. First, we extracted the 95\% confidence interval and the point estimate effect for each study. These values are typically presented in the format:
\[
\text{Point Estimate} \; [\text{Lower Bound}, \text{Upper Bound}]
\]
Next, we identified the appropriate threshold value for comparison. For ratios, this threshold is \textbf{1}, as values greater than 1 indicate a favouring of the intervention and values less than 1 favour the control. For mean differences, the threshold is \textbf{0}, with positive values suggesting favouritism towards the intervention group and negative values favouring the control.

The final step involved comparing the confidence interval to the threshold to determine the conclusion. If the entire confidence interval lies entirely below the threshold (for example, [0.2, 0.8] with a threshold of 1), the study is interpreted as supporting the \textbf{intervention group}. Conversely, if the entire confidence interval is above the threshold (e.g., [1.2, 1.8]), the study supports the \textbf{control group}. If the confidence interval crosses the threshold (e.g., [0.8, 1.2]), the result is not statistically significant, meaning the study supports the \textbf{null hypothesis}.

\begin{table*}
	\centering
		\scalebox{0.8}{
	\begin{subtable}[t]{0.45\textwidth}
		\centering
		\begin{tabular}{lcccc}
			\toprule
			& \textbf{A} & \textbf{B} & \textbf{C} & \textbf{D}  \\
			\midrule
			\textbf{A} & \cellcolor[gray]{0.8}  & 0.94 & 0.93 & 0.95 \\
			\textbf{B} & 0.94 & \cellcolor[gray]{0.8}  & 0.98 & 0.94 \\
			\textbf{C} & 0.93 & 0.98 & \cellcolor[gray]{0.8} & 0.93 \\
			\textbf{D} & 0.95 & 0.94 & 0.93 & \cellcolor[gray]{0.8}  \\
			\bottomrule
		\end{tabular}
		\caption{}
	\end{subtable}
	\hfill
	\begin{subtable}[t]{0.45\textwidth}
		\centering
		\begin{tabular}{lcccc}
			\toprule
			& \textbf{A} & \textbf{B} & \textbf{C} & \textbf{D} \\
			\midrule
			\textbf{A} & \cellcolor[gray]{0.8}  &   0.90    &  0.82     &  0.89     \\
			\textbf{B} &  0.90       & \cellcolor[gray]{0.8}  &    0.92   &    0.97\\ 
			\textbf{C} &  0.82     &  0.92     & \cellcolor[gray]{0.8} &  0.91      \\
			\textbf{D} &   0.89    &    0.97   &      0.91 & \cellcolor[gray]{0.8}  \\
			\bottomrule
		\end{tabular}
		\caption{}
	\end{subtable}
}
	\caption{Pairwise cosine similarity $s(a_i, a_j)$ between annotators for Task 1(a) and Task 3 (b).}
	\label{tab:pairwise-cos-sim}
\end{table*}

\begin{table*}
	\centering
			\scalebox{0.8}{
	\begin{subtable}[t]{0.45\textwidth}
		\centering
		\begin{tabular}{lcccc}
			\toprule
			& \textbf{A} & \textbf{B} & \textbf{C} & \textbf{D} \\
			\midrule
			\textbf{A} & \cellcolor[gray]{0.8}  & 1.00 & 1.00 & 0.87 \\
			\textbf{B} & 1.00 & \cellcolor[gray]{0.8}  & 0.87 & 1.00 \\
			\textbf{C} & 1.00 & 0.87 & \cellcolor[gray]{0.8} & 1.00 \\
			\textbf{D} & 0.87 & 1.00 & 1.00 & \cellcolor[gray]{0.8}  \\
			\bottomrule
		\end{tabular}
		\caption{}
	\end{subtable}
	\hfill
	\begin{subtable}[t]{0.45\textwidth}
		\centering
		\begin{tabular}{lcccc}
			\toprule
			& \textbf{A} & \textbf{B} & \textbf{C} & \textbf{D} \\
			\midrule
			\textbf{A} & \cellcolor[gray]{0.8}  &   0.53    &  0.6     &  0.53     \\
			\textbf{B} &  0.53      & \cellcolor[gray]{0.8}  &    0.6  &    0.6\\ 
			\textbf{C} &  0.6     &  0.6     & \cellcolor[gray]{0.8} &  0.67      \\
			\textbf{D} &   0.53    &    0.6   &      0.67 & \cellcolor[gray]{0.8}  \\
			\bottomrule
		\end{tabular}
		\caption{}
	\end{subtable}
}
	\caption{Pairwise $\varphi$ between annotators for Task 1 (a) and Task 3 (b).}
	\label{tab:pairwise-frac}
\end{table*}

\begin{table*}
	\centering
			\scalebox{0.8}{
	\begin{subtable}[t]{0.45\textwidth}
		\centering
		\begin{tabular}{lcccc}
			\toprule
			& \textbf{A} & \textbf{B} & \textbf{C} & \textbf{D} \\
			\midrule
			\textbf{A} & \cellcolor[gray]{0.8}  & 51.27 & 63.53 & 33.27 \\
			\textbf{B} & 51.27 & \cellcolor[gray]{0.8}  & 30.53 & 51.33 \\
			\textbf{C} & 63.53 & 30.53 & \cellcolor[gray]{0.8} & 60.53 \\
			\textbf{D} & 33.27 & 51.33 & 60.53 & \cellcolor[gray]{0.8}  \\
			\bottomrule
		\end{tabular}
		\caption{}
	\end{subtable}
	\hfill
	\begin{subtable}[t]{0.45\textwidth}
		\centering
		\begin{tabular}{lcccc}
			\toprule
			& \textbf{A} & \textbf{B} & \textbf{C} & \textbf{D} \\
			\midrule
			\textbf{A} & \cellcolor[gray]{0.8}  &   15.67    &  17.87     &  16.73     \\
			\textbf{B} &  15.67      & \cellcolor[gray]{0.8}  &    9.60  &    4.33\\ 
			\textbf{C} &  17.87    &  9.60     & \cellcolor[gray]{0.8} &  8.47      \\
			\textbf{D} &   16.73    &    4.33   &      8.47 & \cellcolor[gray]{0.8}  \\
			\bottomrule
		\end{tabular}
		\caption{}
	\end{subtable}
}
	\caption{Pairwise $lev(a_i, a_j)$ between annotators for Task 1 (a) and Task 3 (b).}
	\label{tab:pairwise-lev}
\end{table*}

 \begin{table*}
 	\centering
 	\scalebox{0.8}{
 	\begin{subtable}[t]{0.45\textwidth}
 		\centering
 		\begin{tabular}{lcccc}
 			\toprule
 			& \textbf{A} & \textbf{B} & \textbf{C} & \textbf{D} \\
 			\midrule
 			\textbf{A} & \cellcolor[gray]{0.8}  & 0.47 & 0.36 & 0.19 \\
 			\textbf{B} & 0.47 & \cellcolor[gray]{0.8}  & 0.19 & 0.39 \\
 			\textbf{C} & 0.36 & 0.19 & \cellcolor[gray]{0.8} & 0.43 \\
 			\textbf{D} & 0.19 & 0.39 & 0.43 & \cellcolor[gray]{0.8}  \\
 			\bottomrule
 		\end{tabular}
 		\caption{}
 	\end{subtable}
 	\hfill
 	\begin{subtable}[t]{0.45\textwidth}
 		\centering
 		\begin{tabular}{lcccc}
 			\toprule
 			& \textbf{A} & \textbf{B} & \textbf{C} & \textbf{D} \\
 			\midrule
 			\textbf{A} & \cellcolor[gray]{0.8}  &   0.62    &  0.42     &  0.30     \\
 			\textbf{B} &  0.62    & \cellcolor[gray]{0.8}  &    0.26  &    0.22\\ 
 			\textbf{C} &  0.42   &  0.26     & \cellcolor[gray]{0.8} &  0.31      \\
 			\textbf{D} &  0.30   &  0.22   &      0.31& \cellcolor[gray]{0.8}  \\
 			\bottomrule
 		\end{tabular}
 		\caption{}
 	\end{subtable}
 }
 	\caption{Pairwise $\text{HTER}(a_i, a_j)$ between annotators for Task 1 (a) and Task 3 (b).}
 	\label{tab:pairwise-hter}
 \end{table*}

 \begin{table*}
	\centering
	\scalebox{0.8}{
	\begin{tabular}{lccc}
		\toprule
		\textbf{Model} & \textbf{\textit{Intervention}} & \textbf{\textit{Inconclusive}} & \textbf{\textit{Control}} \\
		 & \textbf{(F1/P/R)} & \textbf{(F1/P/R)} &\textbf{ (F1/P/R) }\\
		\midrule
		Llama-3.1-70B & 57.4/49.3/68.6 & 65.3/77.1/56.6  & 72.0/65.5/80.0 \\ \hline
		Mistral-Large-2407 &63.1/58.3/68.7  & 68.6/76.9/62.0 & 67.9/60.4/77.8 \\ \hline
		GPT-3.5-turbo-0613 & 54.4/53.9/54.9 & 67.9/72.7/63.7& 64.1/56.9/73.3\\ \hline
		GPT-4-0613 & 61.3/56.6/66.7& 63.9/76.5/54.9& 67.3/55.9/84.5\\
		\bottomrule
	\end{tabular}
}
	\caption{URCA performance metrics by conclusion category.}
	\label{tab:eval-by-label}
\end{table*}

\section{Addition results}\label{appendix:additional-res}

 \subsection{Additional evaluations for URCA}
We report precision, recall and F1 score for type of conclusion (\textit{Intervention}, \textit{Control}, \textit{Inconclusive}) in Table~\ref{tab:eval-by-label}. 

\subsection{Qualitative Example}
\label{appendix:example}
Figure~\ref{fig:qualitative-example} presents a real example from our dataset \textsc{CochraneForest}. URCA semantically clusters the retrieved chunks into three groups: the first one contains relevant information on the outcome under assessment (\textit{fistula closure}), the second one focuses on numerical aspects related to the time to closure and systemic markets, and the third one on a different outcome (\textit{Crohn’s Disease Active Index}). Coloured text indicates the extracted information. By grouping and filtering chunks within clusters, irrelevant information is discarded early. In addition, clustering keeps related chunks together, allowing for better judgment of ambiguous or mixed data (e.g., in Cluster 2). In contrast, without clustering, the LLM doesn't properly handle the context and extracts information regarding the non-statistical significance of PDAI and CDI ("There was no difference in Crohn’s Disease Activity Index (CDAI) score between stem cell transplantation and placebo", "[...] we noted no significant differences between treatment groups"), thus incorrectly concluding the overall study is inconclusive with respect to the research question.

\section{Prompt templates}\label{appendix:prompts}
We provide the prompts we used to execute all our experiments in Figure~\ref{fig:prompts}. 

\section{Annotation Interface}\label{appendix:annotation_interface}
The annotation interface used to annotate \textsc{CochraneForest} is shown in Figure~\ref{fig:annot-ui}. Specifically, Figure~\ref{fig:annot-overview} presents an overview of the interface, which contains a visualisation of the forest plot, title and abstract of the systematic review, as well as links to the Cochrane review; Figure~\ref{fig:annot-tasks} presents the annotation tasks; Figure~\ref{fig:annot-example} illustrates an annotation example for Task 1 and Task 3.

    \section{Hyperparameters and APIs}
    We executed all the experiments either via API or on our own cluster. We used the paid-for OpenAI API to access GPT-3.5.-turbo and GPT-4. On the other hand, we hosted the open-source models used in this paper on a distributed cluster containing  a total of 176 NVIDIA H100 GPUs and served them with vllm~\cite{vllm}. As explained in Section~\ref{rag-settings}, we set the temperature to 0 and the maximum number of tokens to 1,024 for all models. We left all the other hyperparameters to the default value.

     \section{Scientific artefacts and licensing}
    In this work, we used the following scientific artefacts. LLaMa 3.1 is licensed under a commercial license.\footnote{\url{https://llama.meta.com/doc/overview}} GPT-3.5 Turbo and GPT-4 are licensed under a commercial license.\footnote{\url{https://openai.com/policies/terms-of-use/}} 
    Mistral Large is licensed under the Mistral Research License.\footnote{\url{https://mistral.ai/licenses/MRL-0.1.md}}.
    Mining text and data from the Cochrane library is permitted for non-commercial research through the Wiley API.\footnote{\url{https://www.cochranelibrary.com/help/access}}
    The usage of the listed artefacts is consistent with their licenses.     
    \begin{figure*}
    \centering
    \includegraphics[width=\textwidth]{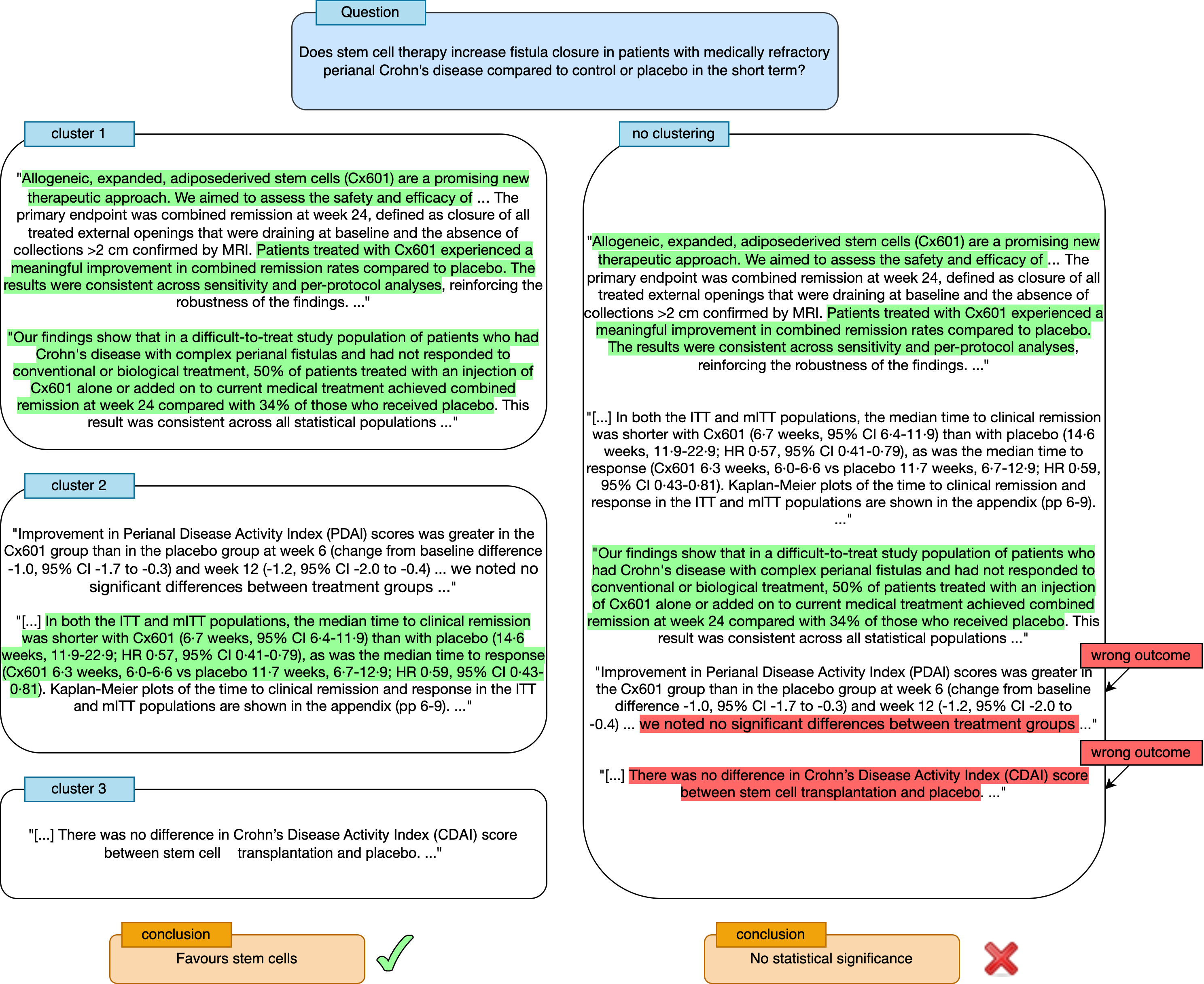}
    \caption{Qualitative example from \textsc{CochraneForest}. URCA clusters the retrieved passages and filters out irrelevant information from each cluster.}
    \label{fig:qualitative-example}
\end{figure*}

    \begin{figure*}
        \centering
        \vspace{-.3cm}
          \begin{subfigure}{.6\textwidth}
     \includegraphics[width=\columnwidth]{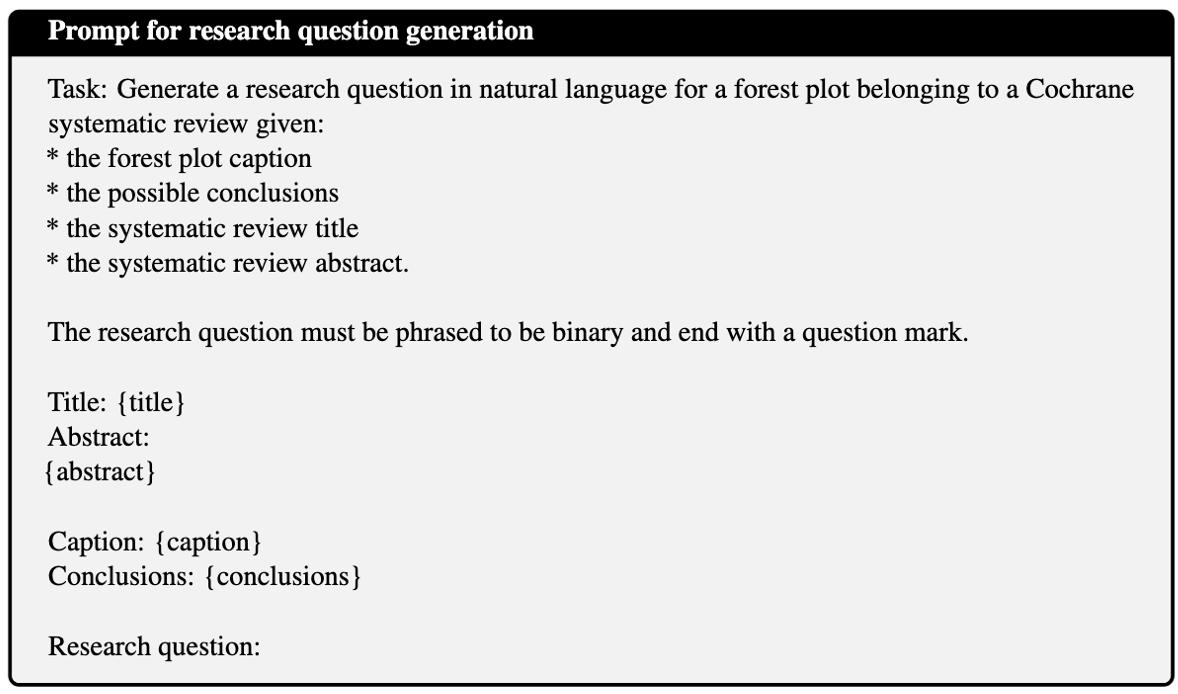}
     \caption{}
     \end{subfigure}
     
     \begin{subfigure}{.6\textwidth}
     \includegraphics[width=\columnwidth]{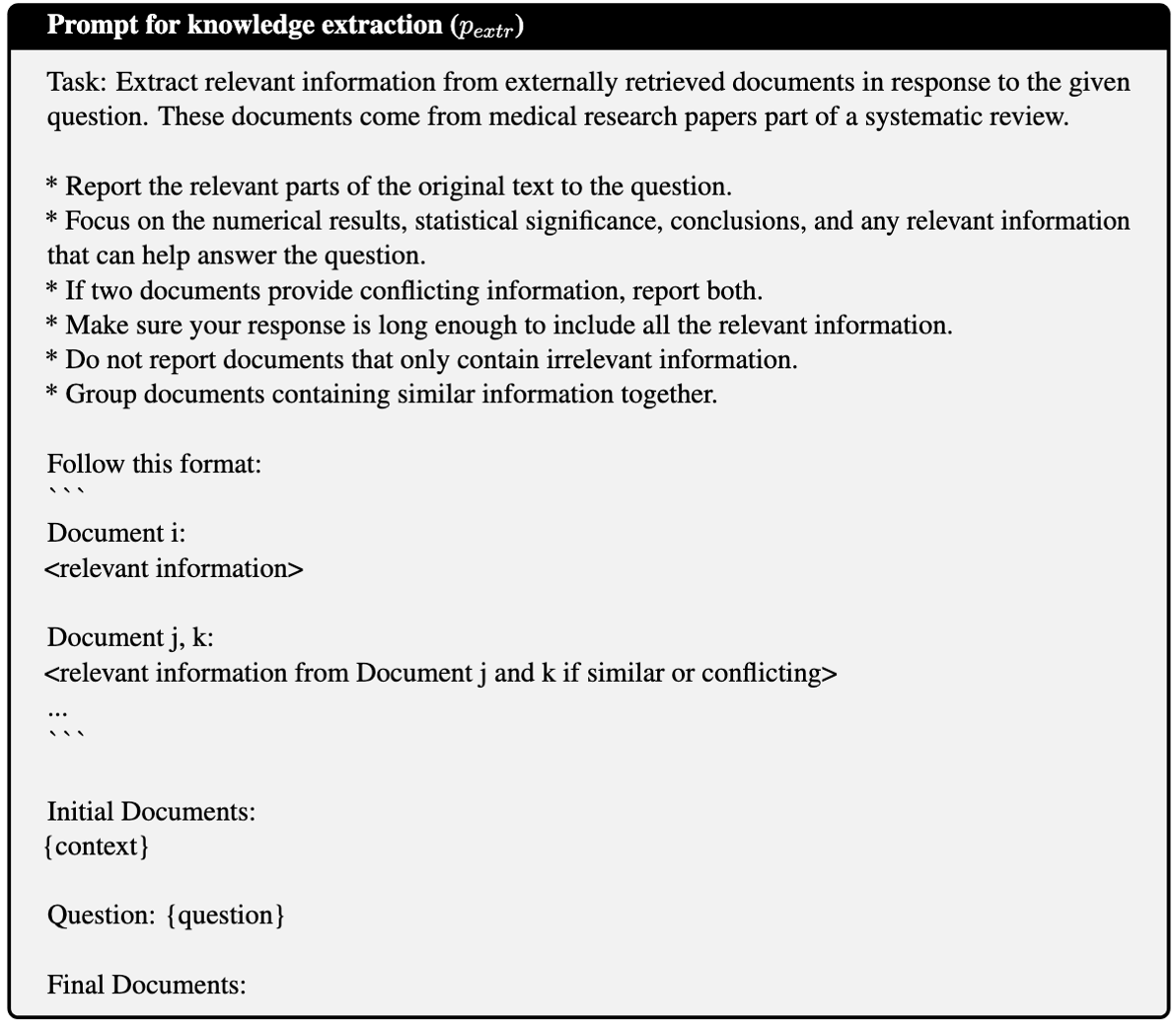}
     \caption{}
     \end{subfigure}

     \begin{subfigure}{.6\textwidth}
     \includegraphics[width=\columnwidth]{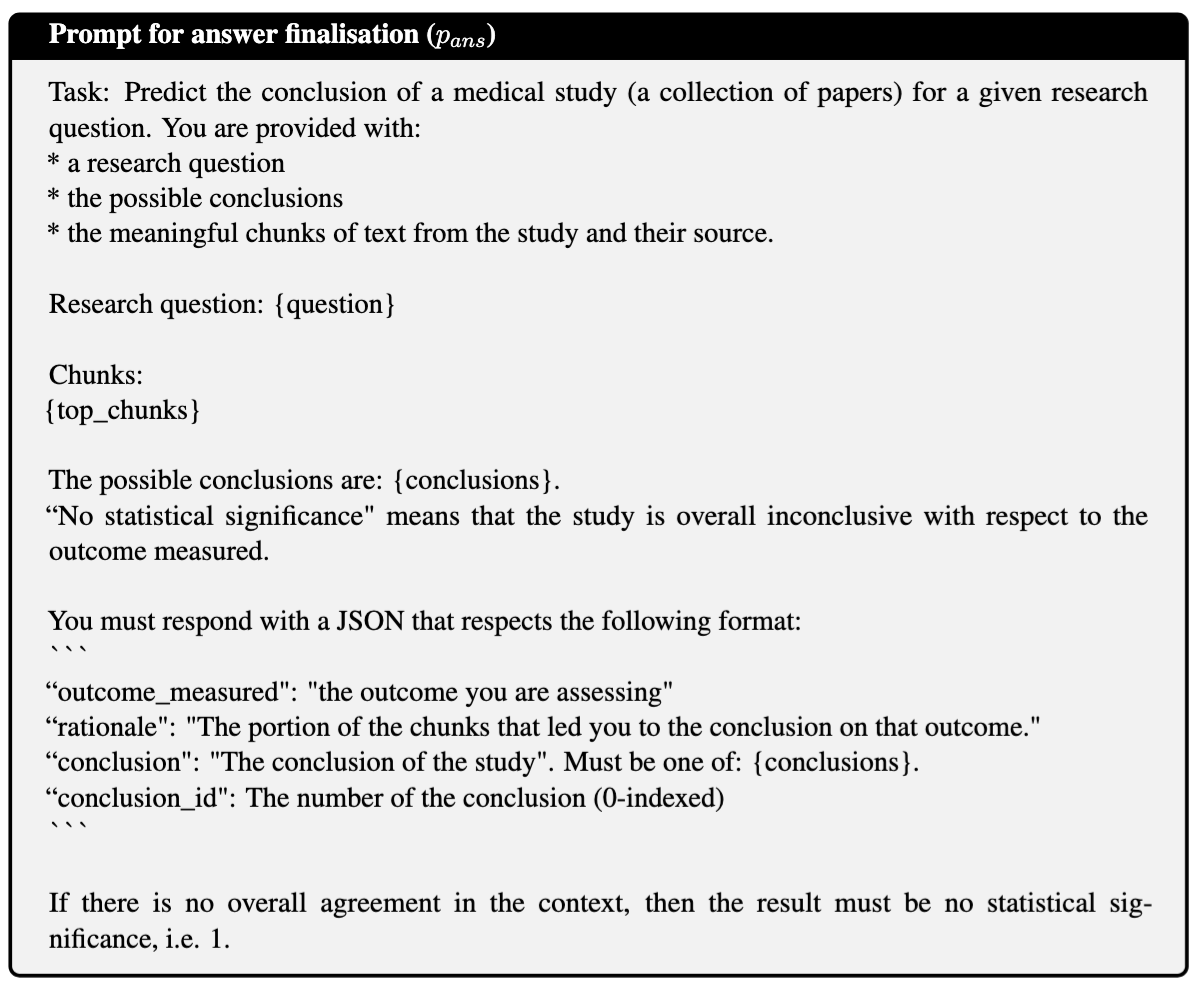}
     \caption{}
     \end{subfigure}

    \caption{Prompt templates for research question generation (a), knowledge extraction (b), and answer finalisation (c).}
    \label{fig:prompts}
    \end{figure*}
    


\begin{figure*}[htbp]
    \centering
    \begin{subfigure}{0.85\textwidth}
        \includegraphics[width=\textwidth]{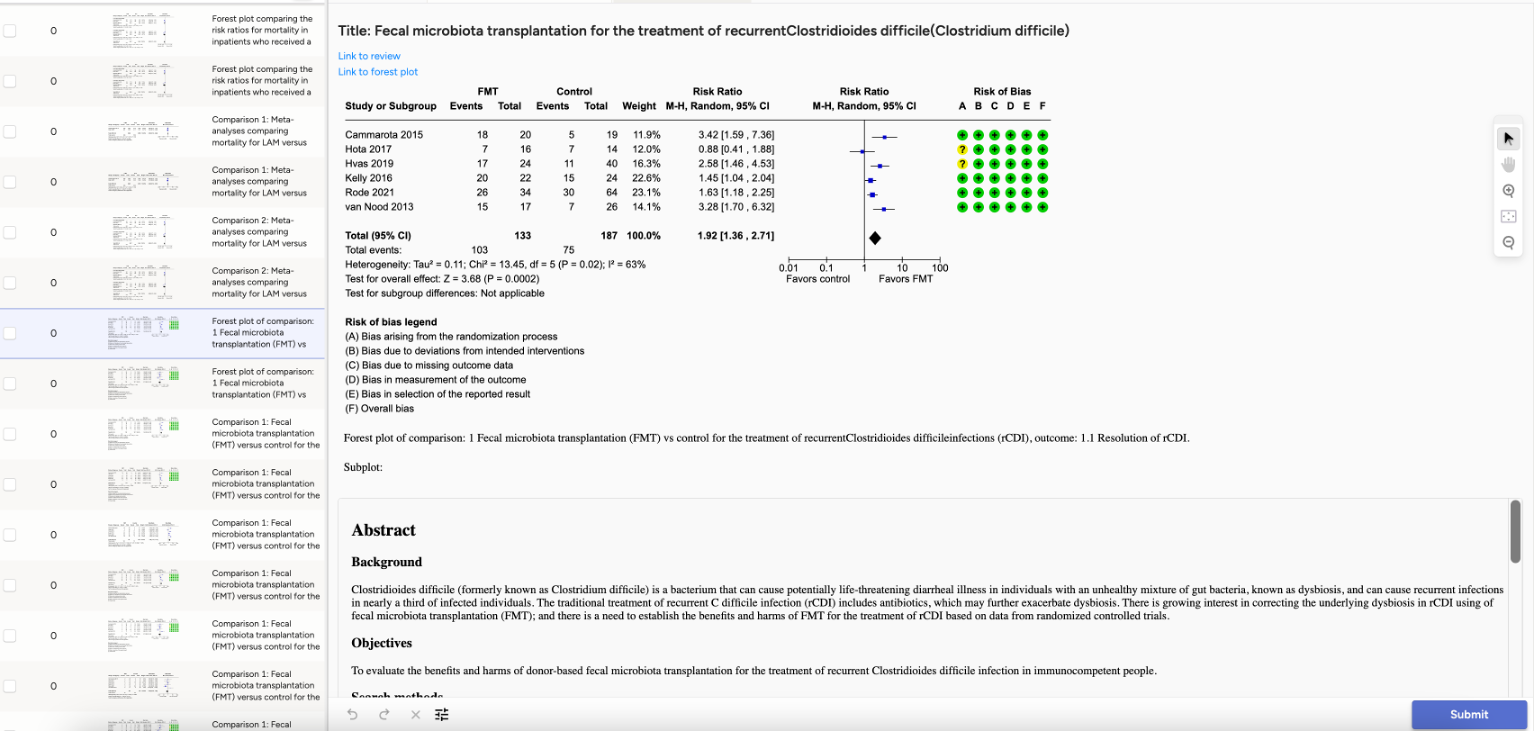} 
        \caption{Main part of the interface, containing a visualisation of the forest plot, title, and long abstract of the systematic review.}
        \label{fig:annot-overview}
    \end{subfigure}
    \hfill
    \begin{subfigure}{0.85\textwidth}
        \includegraphics[width=\textwidth]{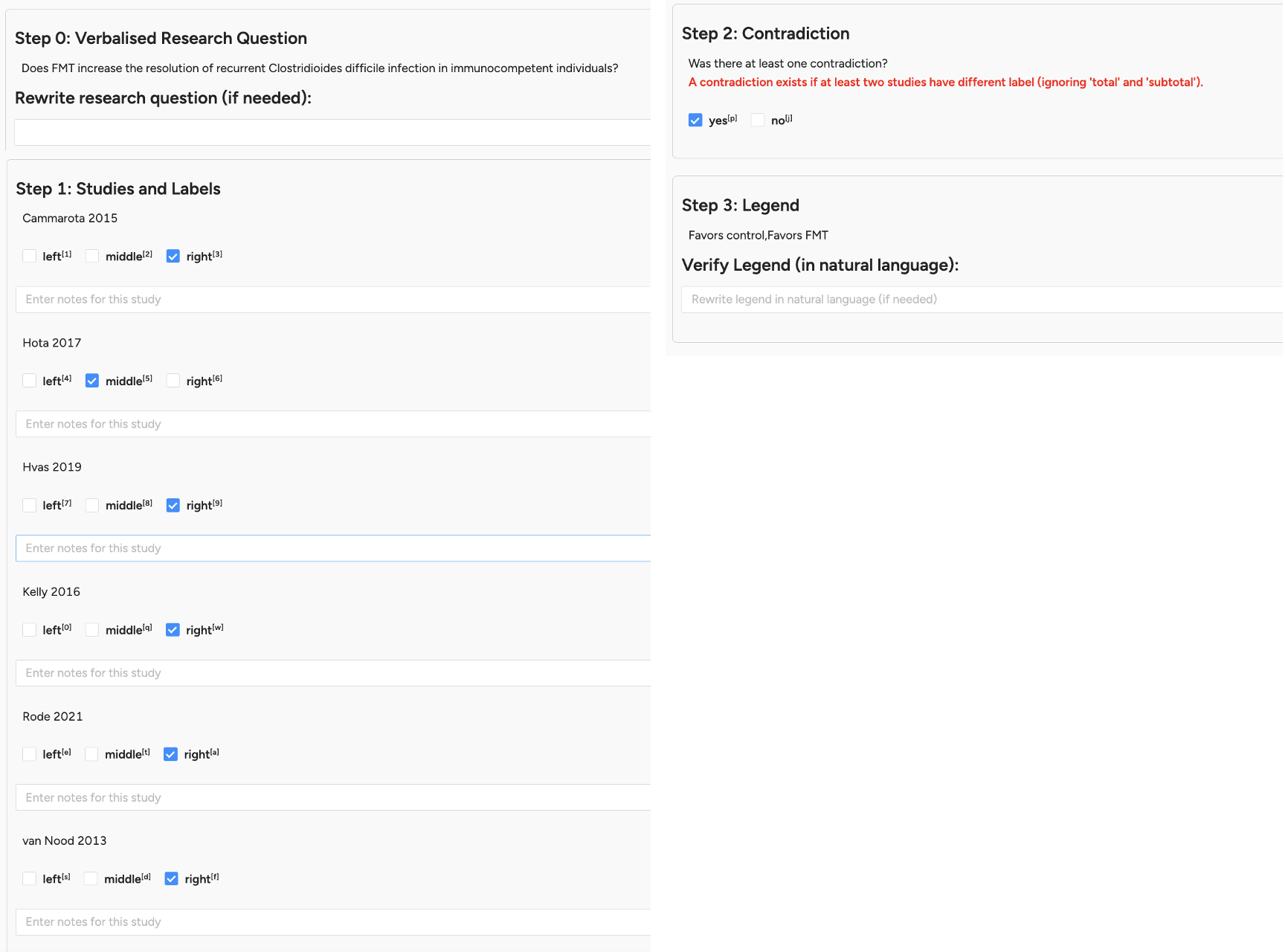} 
        \caption{Annotation tasks.}
        \label{fig:annot-tasks}
    \end{subfigure}
    \hfill
    \begin{subfigure}{0.85\textwidth}
        \includegraphics[width=\textwidth]{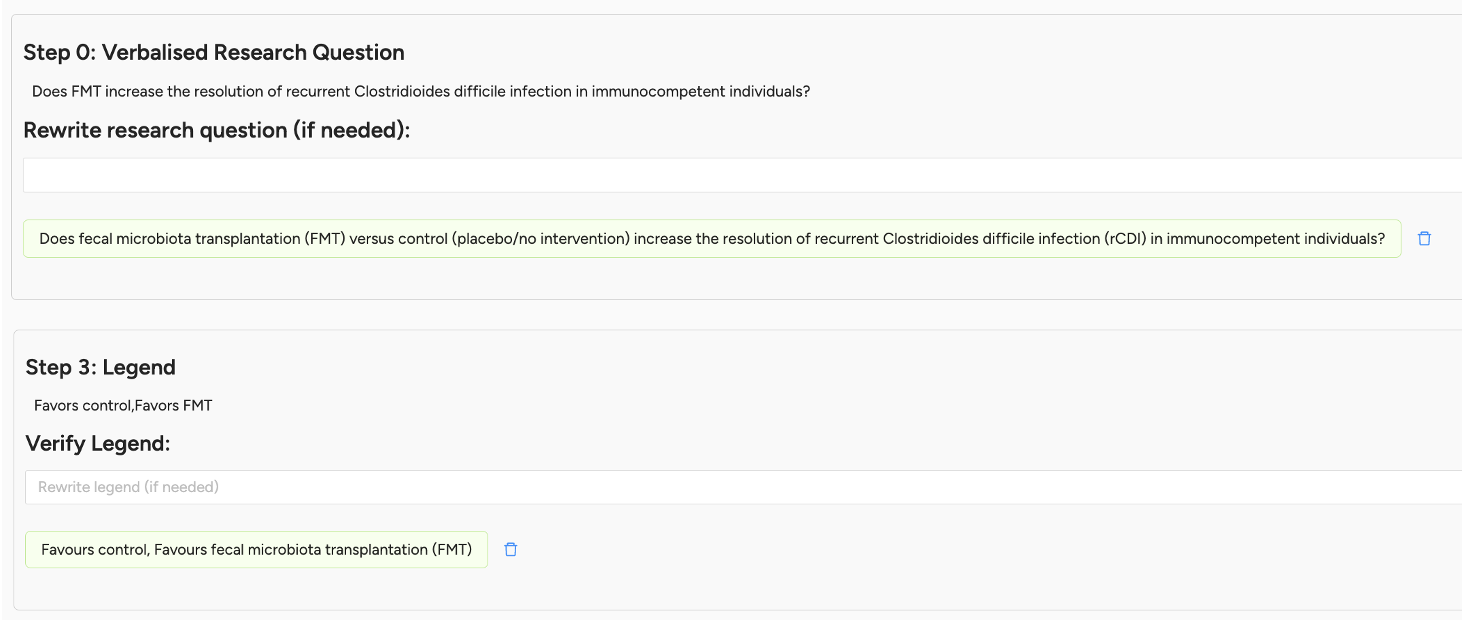} 
        \caption{Example annotation for Task 1 (expanded acronyms and added missing comparator) and Task 3 (expanded acronyms).}
        \label{fig:annot-example}
    \end{subfigure}

    \caption{The forest plot annotation interface. }
    \label{fig:annot-ui}
\end{figure*}

\end{document}